\documentclass[10pt,twocolumn,letterpaper]{article}

\usepackage{iccv}
\usepackage{times}
\usepackage{epsfig}
\usepackage{graphicx}
\usepackage{placeins}
\usepackage{amsmath}
\usepackage{amssymb}
\usepackage{adjustbox}

\usepackage{pifont}
\newcommand{\cmark}{\ding{51}}%
\newcommand{\xmark}{\ding{55}}%

\usepackage{color}
\definecolor{red}{rgb}{0.7,0,0}
\definecolor{green}{rgb}{0.0,0.7,0}
\definecolor{blue}{rgb}{0.00,0.00,0.75}
\definecolor{orange}{rgb}{0.72,0.22,0.06}
\definecolor{purple}{rgb}{0.6,0.0,0.6}
\definecolor{pink}{rgb}{1,0.03,0.5}
\definecolor{olive}{rgb}{0.4,0.6,0}
\definecolor{olive}{rgb}{0,0,0}

\newcommand{\dani}[1]{\textcolor{orange}{(Dani: #1)}}

\newcommand{\iccvNew}[1]{\textcolor{black}{#1}}
\newcommand{\iccvOut}[1]{}

\usepackage{bm}

\definecolor{ticgreen}{rgb}{0.55, 0.71, 0.0}

\usepackage[ruled]{algorithm2e} %
\usepackage{multirow}
\usepackage{colortbl}
\usepackage{color}

\usepackage[pagebackref=true,breaklinks=true,letterpaper=true,colorlinks,bookmarks=false]{hyperref}

\iccvfinalcopy %

\ificcvfinal\fi

\begin{document}

\title{ScanGAN360: A Generative Model of Realistic Scanpaths for 360$^\circ$ Images}

\author{Daniel Martin$^{1}$
\and
Ana Serrano$^{2}$
\and
Alexander W. Bergman$^{3}$
\and
Gordon Wetzstein$^{3}$
\and
Belen Masia$^{1}$\\
\\
\small{$^{1}$Universidad de Zaragoza, I3A\hspace{3em}$^{2}$Centro Universitario de la Defensa, Zaragoza\hspace{3em}$^{3}$Stanford University}
}

\maketitle

\ificcvfinal\thispagestyle{empty}\fi

\begin{abstract}
   \iccvNew{Understanding and modeling the dynamics of human gaze behavior in 360$^\circ$ environments is a key challenge in computer vision and virtual reality. Generative adversarial approaches could alleviate this challenge by generating a large number of possible scanpaths for unseen images.} Existing methods for scanpath generation, however, do not adequately predict realistic scanpaths for 360$^\circ$ images. We \iccvNew{present ScanGAN360,} a new generative adversarial approach to address this challenging problem. Our network generator is tailored to the specifics of \iccvNew{360$^\circ$ images representing immersive environments. Specifically, we accomplish this by leveraging the use of a spherical adaptation of dynamic-time warping as a loss function and proposing a novel parameterization of 360$^\circ$ scanpaths.} The quality of our scanpaths outperforms competing approaches by a large margin and is almost on par with the human baseline. ScanGAN360 thus allows fast simulation of large numbers of \emph{virtual observers}, whose behavior mimics real users, \iccvNew{enabling a better understanding of gaze behavior and novel applications in virtual scene design}.

\end{abstract}

\section{Introduction}
\label{sec:intro}

Virtual reality (VR) is an emerging medium that unlocks unprecedented user experiences. To optimize these experiences, however, it is crucial \iccvNew{to develop computer vision techniques that help us} understand how people explore immersive virtual environments. Models for time-dependent visual exploration behavior are important for designing and editing VR content~\cite{Serrano_VR-cine_SIGGRAPH2017}, for generating realistic gaze trajectories of digital avatars \cite{horley2004face}, for understanding dynamic visual attention and visual search behavior~\cite{yun2013exploring}, and for developing new rendering, display, and compression algorithms, among other applications.

Current approaches that model how people explore virtual environments often leverage  saliency prediction~\cite{sitzmann2018saliency,coors2018spherenet,martin20saliency,assens2017saltinet}. While this is useful for some applications, the fixation points predicted by these approaches do not account for the time-dependent visual behavior of the user, making it difficult to predict the order of fixations, or give insight into how people explore an environment over time. For this purpose, some recent work has explored scanpath prediction~\cite{assens2017saltinet,assens2018pathgan,zhu2018prediction,assens2018scanpath}, but these algorithms 
do not adequately model how people explore immersive virtual environments, resulting in erratic or non-plausible scanpaths.

\begin{figure}[t]
	\centering
	\includegraphics[width=\linewidth]{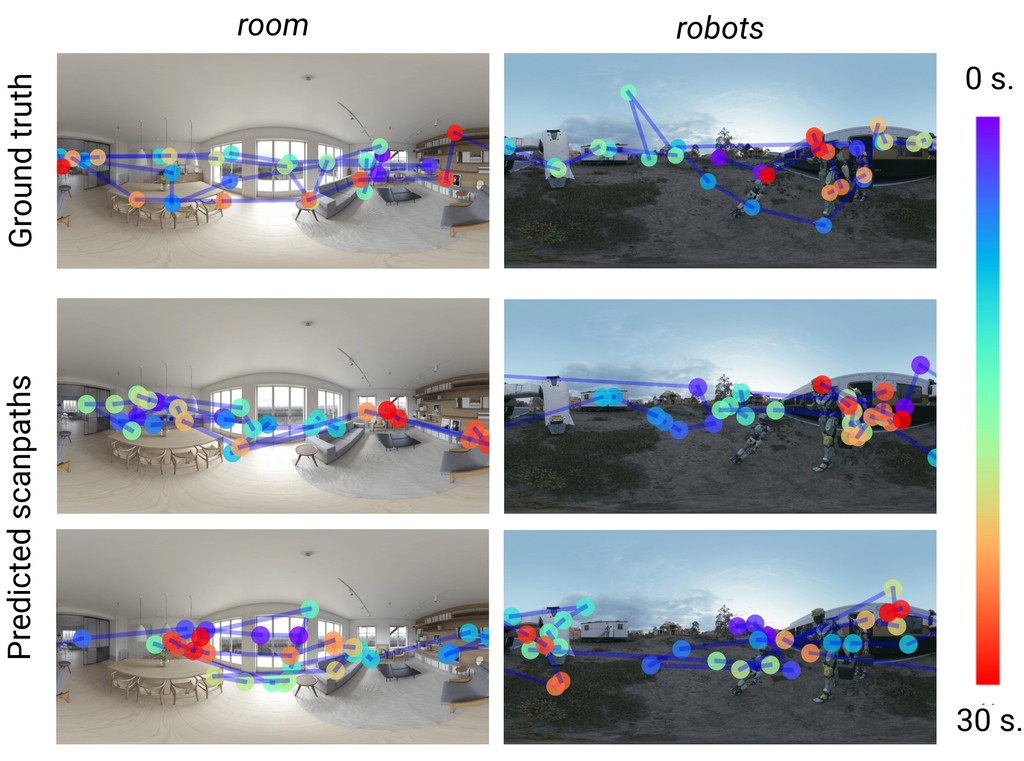}
	\caption{\iccvNew{We present \iccvNew{ScanGAN360,} a generative adversarial approach to scanpath generation for 360$^{\circ}$ images. ScanGAN360 generates realistic scanpaths (\emph{bottom rows}), outperforming state-of-the-art methods and mimicking the human baseline (\emph{top row}).}}

	\label{fig:teaser}
\end{figure}

In this work, we present \iccvNew{ScanGAN360,} a novel framework for scanpath generation for 360$^\circ$ images (Figure~\ref{fig:teaser}). 
Our model builds on a conditional generative adversarial network (cGAN) architecture,  for which we discuss and validate two important insights that we show are necessary for realistic scanpath generation.
First, we propose a loss function based on a spherical adaptation of dynamic time warping (DTW), which is a key aspect for training our GAN robustly. 
DTW is a metric for measuring similarity between two time series, such as scanpaths, which to our knowledge has not been used to train scanpath-generating GANs. 
Second, to adequately tackle the problem of scanpath generation in 360$^{\circ}$ images, we present a novel parameterization of the scanpaths. 
These insights allow us to demonstrate state-of-the-art results for scanpath generation in VR, close to the human baseline and far surpassing the performance of existing methods. Our approach is the first to enable robust scanpath prediction over long time periods up to 30 seconds, and, unlike previous work, our model does not rely on saliency, which is typically not available as ground truth.

Our model produces about 1,000 scanpaths per second, which enables fast simulation of large numbers of \emph{virtual observers}, whose behavior mimics that of real users. %
\iccvNew{Using ScanGAN360, we explore applications in virtual scene design, which is useful in video games, interior design, cinematography, and tourism, and scanpath-driven video thumbnail generation of 360$^\circ$ images, which provides previews of VR content for social media platforms. Beyond these applications, we propose to use ScanGAN360 for applications such as gaze behavior simulation for virtual avatars or gaze-contingent rendering. Extended discussion and results on applications are included in the supplementary material and video.}

We will make our source code and pre-trained model publicly available to promote future research.

\section{Related work}
\label{sec:relatedwork}

\paragraph{Modeling and predicting attention} 
The multimodal nature of attention~\cite{martin2021multimodality}, together with the complexity of human gaze behavior, make this a very challenging task. Many works devoted to it have relied on representations such as saliency, which is a convenient representation for indicating the regions of an image more likely to attract attention. Early strategies for saliency modeling have focused on either creating hand-crafted features representative of saliency~\cite{itti1998model,saliencytoolbox, zhao2011saliency, lu2012cvpr, judd2009learning, borji2012cvpr}, or directly learning data-driven features~\cite{torralba2006contextual, kummerer2016deepgaze}. With the proliferation of extensive datasets of human attention~\cite{sitzmann2018saliency, rai2017dataset, judd2009learning, mit-saliency-benchmark, yang2013saliency}, deep learning--based methods for saliency prediction have been successfully applied, yielding impressive results~\cite{Pan_2016_CVPR, Pan_2017_SalGAN, cornia2018predicting, Vig_2014_CVPR, wang2018attention, Wang_2018_CVPR, xu2019pami}. 

However, saliency models do not take into account the dynamic nature of human gaze behavior, and therefore, they are unable to model or predict time-varying aspects of attention. 
Being able to model and predict dynamic exploration patterns has been proven to be useful, for example, for avatar gaze control~\cite{colburn2000role, sela2017gazegan}, video rendering in virtual reality~\cite{ling2019prediction}, or for directing users' attention over time in many contexts~\cite{cao2014look, pang2016directing}.
Scanpath models aim to predict visual patterns of exploration that an observer would perform when presented with an image. In contrast to  saliency models, scanpath models typically focus on predicting plausible scanpaths, \ie, they do not predict a unique scanpath and instead they try to mimic human behavior when exploring an image, taking into account the variability between different observers. Ellis and Smith~\cite{ellis1985patterns} were pioneers in this field: they proposed a general framework for generating scanpaths based on Markov stochastic processes. Several approaches have followed this work, incorporating behavioral biases in the process in order to produce more plausible scanpaths~\cite{lemeur2015saccadic, tatler2009prominence,liu2013semantically,tavakoli2013stochastic}. In recent years, deep learning models have been used to predict human scanpaths based on neural network features trained on object recognition~\cite{kummerer2016deepgaze, wang2017deep, cornia2018predicting, bao2020scanpath}. 

\paragraph{Attention in 360$^\circ$ images} 
Predicting plausible scanpaths in 360$^\circ$ imagery is a more complex task: Observers do not only scan a given image with their gaze, but they can now also turn their head or body, effectively changing their viewport over time. Several works have been proposed for modeling saliency in 360$^\circ$ images~\cite{monroy2018salnet, sitzmann2018saliency,martin20saliency,chao2018salgan360,startsev2018360}. 
However, scanpath prediction has received less attention. In their recent work, Assens et al.~\cite{assens2018pathgan} generalize their 2D model to 360$^\circ$ images, but their loss function is unable to reproduce the behavior of ground truth scanpaths (see Figure~\ref{fig:comparison}, third column).
A few works have focused on predicting short-term sequential gaze points based on users' previous history for 360$^\circ$ videos, but they are limited to small temporal windows (from one to ten seconds)~\cite{wu2020spherical, li2019very, nguyen2018your}.
For the case of images, a number of recent methods focus on developing improved saliency models and principled methods to sample from them~\cite{assens2017saltinet, assens2018scanpath, zhu2018prediction}.

Instead, we directly learn dynamic aspects of attention from ground truth scanpaths by training a generative model in an adversarial manner, with an architecture and loss function specifically designed for scanpaths in 360$^\circ$ images.
This allows us to (i) effectively mimic human behavior when exploring scenes, bypassing the saliency generation and sampling steps, and (ii) optimize our network to stochastically generate 360$^\circ$ scanpaths, taking into account observer variability.

\section{Our Model}
\label{sec:model}

We adopt a generative adversarial approach, 
specifically designed for 360$^\circ$ content in which the model learns to generate a plausible scanpath, given the 360$^\circ$ image as a condition. In the following, we describe the parameterization employed for the scanpaths, the design of our loss function for the generator, and the particularities of our conditional GAN architecture, ending with details about the training process.

\begin{figure*}[htb]
	\centering
	\includegraphics[width=\linewidth]{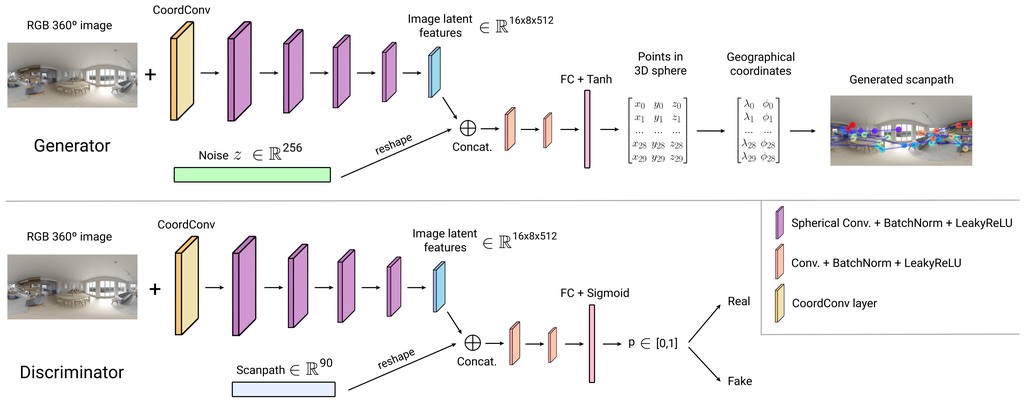}
	\caption{Illustration of our generator and discriminator networks. Both networks have a two-branch structure: Features extracted from the 360$^{\circ}$ image with the aid of a CoordConv layer and an encoder-like network are concatenated with the input vector for further processing. The generator learns to transform this input vector, conditioned by the image, into a plausible scanpath. The discriminator takes as input vector a scanpath (either captured or synthesized by the generator), as well as the corresponding image, and determines the probability of this scanpath being real (or fake). We train them end-to-end in an adversarial manner, following a conditional GAN scheme. Please refer to the text for details on the loss functions and architecture. 
	}
	\label{fig:models}
\end{figure*}

\subsection{Scanpath Parameterization}
\label{subsec:sph_param}

Scanpaths are commonly provided as a sequence of two-dimensional values corresponding to the coordinates $(i,j)$ of each gaze point in the image. When dealing with 360$^\circ$ images in equirectangular projections, gaze points are also often represented by their latitude and longitude $(\phi, \lambda)$, $\phi \in [-\frac{\pi}{2}, \frac{\pi}{2}]$ and $\lambda \in [-\pi,\pi]$. However, these parameterizations either suffer from discontinuities at the borders of a 360$^\circ$ image, or result in periodic, ambiguous values. The same point of the scene can have two different representations in these parameterizations, hindering the learning process. 

We therefore resort to a three-dimensional parameterization of our scanpaths, where each gaze point ${p} = (\phi, \lambda)$ is transformed into its three-dimensional representation $P=(x,y,z)$ such that:
\begin{equation*}
    x = \text{cos}(\phi)\,\text{cos}(\lambda); \,
    y = \text{cos}(\phi)\,\text{sin}(\lambda); \,
    z = \text{sin}(\phi).
    \label{eq:to_3d}
\end{equation*}
This transformation assumes, without loss of generality, that the panorama is projected over a unit sphere. We use this parameterization for our model, which learns a scanpath $\mathbf{P}$ as a set of three-dimensional points over time. Specifically, given a number of samples $T$ over time, $\mathbf{P}=(P_1,...,P_T)\in \mathbb{R}^{3 \times T}$.
The results of the model are then converted back to a two-dimensional parameterization in terms of latitude ($\phi = \text{atan2}(z, \sqrt{x^2 + y^2})$) and longitude ($\lambda = \text{atan2}(y, x)$) for display and evaluation purposes.

\subsection{Overview of the Model}
\label{subsec:overview}

Our model is a conditional GAN, where the condition is the RGB 360$^\circ$ image for which we wish to estimate a scanpath. 
The generator $G$ is trained to generate a scanpath from a latent code $z$ (drawn randomly from a uniform distribution, $\mathcal{U}(-1,1)$), conditioned by the RGB 360$^\circ$ image $y$. %
The discriminator $D$ takes as input a potential scanpath ($x$ or $G(z,y)$), as well as the condition $y$ (the RGB 360$^\circ$ image), and outputs the probability of the scanpath being real (or fake).
The architecture of both networks, generator and discriminator, can be seen in Figure~\ref{fig:models}, and further details related to the architecture are described in Section~\ref{subsec:arch}.

\subsection{Loss Function}
\label{subsec:loss}

The objective function of a conventional conditional GAN is inspired by a minimax objective from game theory, with an objective~\cite{mirza2014conditional}: 
\begin{equation}
\begin{split}
& \min_G \max_D V(D,G) = \\ 
& \mathbb{E}_x [\log D(x,y) ] + \mathbb{E}_z [\log (1-D(G(z,y),y)].
\end{split}
\end{equation}
We can separate this into two losses, one for the generator, $\mathcal{L}_G$, and one for the discriminator, $\mathcal{L}_D$: 
\begin{equation}
\mathcal{L}_{G} = \mathbb{E}_z [\log (1-D(G(z,y),y)) ] ,
\label{eq:loss_gen}
\end{equation}
\begin{equation}
\mathcal{L}_{D} = \mathbb{E}_x [ \log D(x,y) ] + \mathbb{E}_z [ \log (1-D(G(z,y),y)) ] .
\label{eq:loss_disc}
\end{equation}

While this objective function suffices in certain cases, as the complexity of the problem increases, the generator may not be able to learn the transformation from the input distribution into the target one. One can resort to adding a loss term to $\mathcal{L}_G$, and in particular one that enforces similarity to the scanpath ground truth data. However, using a conventional data term, such as MSE, does not yield good results (Section~\ref{subsec:model_ablation} includes an evaluation of this). To address this issue, we introduce a novel term in $\mathcal{L}_G$ specifically targeted to our problem, and based on dynamic time warping~\cite{muller2007dynamic}. 

Dynamic time warping (DTW) measures the similarity between two temporal sequences, considering both the shape and the order of the elements of a sequence, without forcing a one-to-one correspondence between elements of the time series. For this purpose, it takes into account all the possible alignments of two time series $\mathbf{r}$ and $\mathbf{s}$, and computes the one that yields the minimal distance between them. Specifically, the DTW loss function between two time series $\mathbf{r} \in \mathbb{R}^{k \times n}$ and $\mathbf{s} \in \mathbb{R}^{k \times m}$ can be expressed as~\cite{cuturi2017soft}:
\begin{equation}
    \textup{DTW}(\mathbf{r}, \mathbf{s}) = \min_A \langle A, \Delta(\mathbf{r}, \mathbf{s}) \rangle,
    \label{eq:dtw_loss}
\end{equation}
where $\Delta(\mathbf{r}, \mathbf{s}) = [\delta(r_i,s_j)]_{ij} \in \mathbb{R}^{n \times m}$ is a matrix containing the distances $\delta(\cdot,\cdot)$ between each pair of points in $\mathbf{r}$ and $\mathbf{s}$, $A$ is a binary matrix that accounts for the alignment (or correspondence) between $\mathbf{r}$ and $\mathbf{s}$, and $\langle\cdot,\cdot\rangle$ is the inner product between both matrices. 

In our case, $\mathbf{r}=(r_1,...,r_T)\in \mathbb{R}^{3 \times T}$ and $\mathbf{s}=(s_1,...,s_T)\in \mathbb{R}^{3 \times T}$
are two scanpaths that we wish to compare. While the Euclidean distance between each pair of points is usually employed when computing $\delta(r_i, s_j)$ for Equation~\ref{eq:dtw_loss}, in our scenario that would yield erroneous distances derived from the projection of the 360$^\circ$ image (both if done in 2D over the image, or in 3D with the parameterization described in Section~\ref{subsec:sph_param}). We instead use the distance over the surface of a sphere, or spherical distance, and define $\Delta_{sph}(\mathbf{r}, \mathbf{s}) = [\delta_{sph}(r_i,s_j)]_{ij} \in \mathbb{R}^{n \times m}$ such that:
\begin{equation}
\begin{split}
    & \delta_{sph}(r_i,s_j) = \\ 
    & 2 \arcsin \left(\frac{1}{2}\sqrt{(r_i^x-s_j^x)^2 + (r_i^y-s_j^y)^2 + (r_{i}^{z}-s_{j}^{z})^2}\right),
    \label{eq:gcd}
\end{split}
\end{equation}
leading to our spherical DTW:
\begin{equation}
    \textup{DTW}_{sph}(\mathbf{r}, \mathbf{s}) = \min_A \langle A, \Delta_{sph}(\mathbf{r}, \mathbf{s}) \rangle .
    \label{eq:sdtw}
\end{equation}
We incorporate the spherical DTW to the loss function of the generator ($\mathcal{L}_G$, Equation~\ref{eq:loss_gen}), yielding our final generator loss function $\mathcal{L}^*_G$:
\begin{equation}
    \mathcal{L}^*_{G} = \mathcal{L}_G + \lambda \cdot \mathbb{E}_z [ \textup{DTW}_{sph}(G(z,y), \mathbf{\rho}) ] ,
\end{equation}
where $\mathbf{\rho}$ is a ground truth scanpath for the conditioning image $y$, and the weight $\lambda$ is empirically set to $0.1$.

While a loss function incorporating DTW (or spherical DTW) is not differentiable, a differentiable version, soft-DTW, has been proposed.
We use this soft-DTW in our model; details on it can be found in Section S1 in the supplementary material or in the original publication~\cite{cuturi2017soft}.

\subsection{Model Architecture}
\label{subsec:arch}

Both our generator and discriminator are based on a two-branch structure (see Figure~\ref{fig:models}), with one branch for the conditioning image $y$ and the other for the input vector ($z$ in the generator, and $x$ or $G(z,y)$ in the discriminator). The image branch extracts features from the 360$^\circ$ image, yielding a set of latent features that will be concatenated with the input vector for further processing. Due to the distortion inherent to equirectangular projections, traditional convolutional feature extraction strategies are not well suited for 360$^\circ$ images: They use a kernel window where neighboring relations are established uniformly around a pixel. Instead, we extract features using panoramic (or spherical) convolutions~\cite{coors2018spherenet}. Spherical convolutions are a type of dilated convolutions where the relations between elements in the image are not established in image space, but in a gnomonic, non-distorted space. These spherical convolutions can represent kernels as patches tangent to a sphere where the 360$^\circ$ is reprojected.

In our problem of scanpath generation, the location of the features in the image is of particular importance. Therefore, to facilitate spatial learning of the network, we use the recently presented CoordConv strategy~\cite{liu2018intriguing}, which gives convolutions access to its own input coordinates by adding extra coordinate channels. We do this by concatenating a CoordConv layer to the input 360$^\circ$ image (see Figure~\ref{fig:models}). 
This layer also helps stabilize the training process, as shown in Section~\ref{subsec:model_ablation}.

\subsection{Dataset and Training Details}
\label{subsec:train}

We train our model using Sitzmann et al.'s \cite{sitzmann2018saliency} dataset, composed of 22 different 360$^{\circ}$ images and a total of 1,980 scanpaths from 169 different users. Each scanpath contains gaze information captured during 30 seconds with a binocular eye tracking recorder at 120~Hz. We sample these captured scanpaths at 1~Hz (\ie, $T=30$), and reparameterize them (Section~\ref{subsec:sph_param}), so that each scanpath is a sequence $\mathbf{P} = (P_0,...,P_{29}) \in \mathbb{R}^{3 \times T}$. Given the relatively small size of the dataset, we perform data augmentation by longitudinally shifting the 360$^{\circ}$ images (and adjusting their scanpaths accordingly); specifically, for each image we generate six different variations with random longitudinal shifting.
We use 19 of the 22 images in this dataset for training, and reserve three to be part of our test set (more details on the full test set are described in Section~\ref{sec:evaluation}). With the data augmentation process, this yields 114 images in the training set.

During our training process we use the Adam optimizer~\cite{Adam}, with constant learning rates $l_G = 10^{-4}$ for the generator, and $l_D = 10^{-5}$ for the discriminator, both of them with momentum = $(0.5, 0.99)$. 
Further training and implementation details can be found in the supplementary material.

\begin{figure*}[th]
	\centering
	\includegraphics[width=\linewidth]{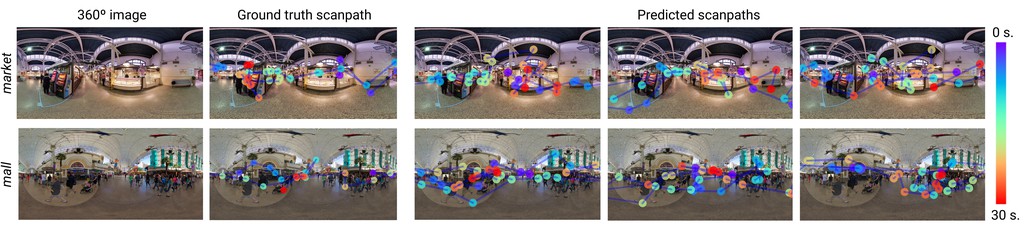}
	\caption{Results of our model for two different scenes: \emph{market} and \emph{mall} from Rai et al.'s dataset~\cite{rai2017dataset}. \emph{From left to right}: 360$^\circ$ image, ground truth sample scanpath, and three scanpaths generated by our model. The generated scanpaths are plausible and focus on relevant parts of the scene, yet they exhibit the diversity expected among different human observers. \iccvNew{Please refer to the supplementary material for a larger set of results.}
	}
	\vspace{0.4em}
	\label{fig:qualitativeresults}
\end{figure*}

\section{Validation and Analysis}
\label{sec:evaluation}

We evaluate the quality of the generated scanpaths with respect to the measured, ground truth scanpaths, as well as to %
other approaches. We also ablate our model to illustrate the contribution of the different design choices. %

We evaluate or model on two different test sets. First, using the three images from Sitzmann et al.'s dataset~\cite{sitzmann2018saliency} left out of the training (Section~\ref{subsec:train}): \emph{room}, \emph{chess} and \emph{robots}. To ensure our model has an ability to extrapolate, we also evaluate it with a different dataset from Rai et al.~\cite{rai2017dataset}. This dataset consists of 60 scenes watched by 40 to 42 observers for 25 seconds. Thus, when comparing to their ground truth, we cut our 30-second scanpaths to the maximum length of their data. 
\iccvNew{Please also refer to the supplementary material for more details on the test set, as well as further evaluation and results.}
\iccvOut{This dataset is larger than Sitzmann et al.'s in size (number of images), but provides gaze data in the form of fixations with associated timestamps, and not the raw gaze points. Note that most of the metrics proposed in the literature for scanpath similarity (Section~\ref{subsec:metrics}) are designed to work with time series of different length, and do not necessarily assume a direct pairwise equivalence, making them valid to compare our generated scanpaths to the ground truth ones from Rai et al.'s dataset. }

\begin{figure*}[t]
	\centering
	\includegraphics[width=\linewidth]{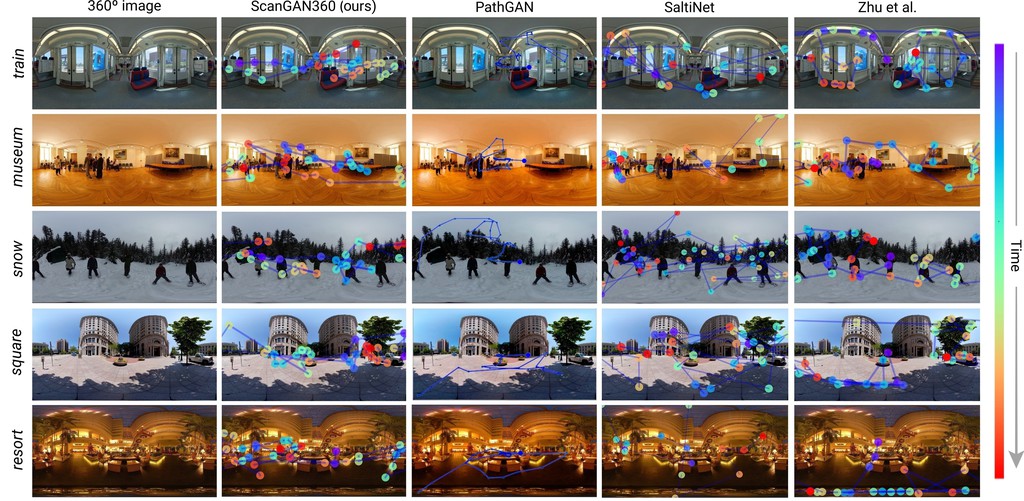}
	\caption{Qualitative comparison to previous methods for five different scenes from Rai et al.'s dataset. In each row, from left to right: 360$^\circ$ image, and a sample scanpath obtained with our method, PathGAN~\cite{assens2018pathgan}, SaltiNet~\cite{assens2018scanpath}, and Zhu et al.'s~\cite{zhu2018prediction}.
	Note that, in the case of PathGAN, we are including the results directly taken from their paper, thus the different visualization.
	Our method produces plausible scanpaths focused on meaningful regions, in comparison with other techniques. Please see text for details, and the supplementary material for a larger set of results, also including ground truth scanpaths. 
	}
	\label{fig:comparison}
\end{figure*}

\subsection{Scanpath Similarity Metrics}
\label{subsec:metrics}

Our evaluation is both quantitative and qualitative. Evaluating scanpath similarity is not a trivial task, and a number of metrics have been proposed in the literature, each focused on a different context or aspect of gaze behavior~\cite{fahimi2020metrics}. Proposed metrics 
can be roughly categorized into: 
(i) direct measures based on Euclidean distance; %
(ii) string-based measures based on string alignment techniques (such as the Levenshtein distance, LEV); %
(iii) curve similarity methods; %
(iv) metrics from time-series analysis (like DTW, on which our loss function is based); and %
(v) metrics from recurrence analysis (\eg, recurrence measure REC and determinism measure DET). %
We refer the reader to \iccvNew{supplementary material and the} review by Fahimi and Bruce~\cite{fahimi2020metrics} for an in-depth explanation and comparison of existing metrics. Here, we include a subset of metrics that take into account both the position and the ordering of the points (namely LEV and DTW), and two metrics from recurrence analysis (REC and DET), which have been reported to be discriminative in revealing viewing behaviors and patterns when comparing scanpaths. %
We nevertheless compute our evaluation for the full set of metrics reviewed by Fahimi and Bruce~\cite{fahimi2020metrics} in the supplementary material.

Since for each image we have a number of ground truth scanpaths, and a set of generated scanpaths, we compute each similarity metric for all possible pairwise comparisons (each generated scanpath against each of the ground truth scanpaths), and average the result. In order to provide an \iccvNew{upper} baseline for each metric, we also compute the human baseline (\textit{Human BL})~\cite{xia2019predicting}, which is obtained by comparing each ground truth scanpath against all the other ground truth ones, and averaging the results. 
\iccvNew{In a similar fashion, we compute a lower baseline based on sampling gaze points randomly over the image (\emph{Random BL}).}

\subsection{Results}
\label{subsec:results}

\iccvNew{Qualitative results of our model can be seen in Figures~\ref{fig:qualitativeresults} and~\ref{fig:teaser} for scenes with different layouts. Figure~\ref{fig:qualitativeresults}, from left to right, shows: the scene, a sample ground truth (captured) scanpath, and three of our generated scanpaths sampled from the generator.} Our model is able to produce plausible, coherent scanpaths that focus on relevant parts of the scene. In the generated scanpaths we observe regions where the user focuses (points of a similar color clustered together), as well as more exploratory behavior. The generated scanpaths are diverse but plausible, as one would expect if different users watched the scene (the supplementary material contains more ground truth, measured scanpaths, showing this diversity). Further, our model is not affected by the inherent distortions of the 360$^\circ$ image. This is apparent, for example, in the \emph{market} scene: The central corridor, narrow and seemingly featureless, is observed by generated \emph{virtual observers}. 
Quantitative results in Table~\ref{tab:metric_results} further show that our generated scanpaths are close to the human baseline (\textit{Human BL}), both in the test set from Sitzmann et al.'s dataset, and over Rai et al.'s dataset. A value close to \textit{Human BL} indicates that the generated scanpaths are as valid or as plausible as the captured, ground truth ones. Note that obtaining a value lower than \textit{Human BL} is possible, if the generated scanpaths are on average closer to the ground truth ones, and exhibit less variance.

Since our model is generative, it can generate as many scanpaths as needed and model many different potential observers. We perform our evaluations on a random set of 100 scanpaths generated by our model. We choose this number to match the number of generated scanpaths available for competing methods, to perform a fair comparison. 
Nevertheless, we have analyzed the stability of our generative model by computing our evaluation metrics for a variable number of generated scanpaths: Our results are very stable with the number of scanpaths (please see Table 2 in the supplementary material).

\subsection{Comparison to Other Methods}
\label{subsec:comparisons}

We compare %
\iccvNew{ScanGAN360} to three methods devoted to scanpath prediction in 360$^\circ$ images: SaltiNet-based scanpath prediction~\cite{assens2017saltinet, assens2018scanpath} (we will refer to it as SaltiNet in the following), PathGAN~\cite{assens2018pathgan} and Zhu et al.'s method~\cite{zhu2018prediction}. For comparisons to SaltiNet we use the public implementation of the authors, while the authors of Zhu et al. kindly provided us with the results of their method for the images from Rai et al.'s dataset (but not for Sitzmann et al.'s); we therefore have both qualitative (Figure~\ref{fig:comparison}) and quantitative (Table~\ref{tab:metric_results}) comparisons to these two methods. In the case of PathGAN, no model or implementation could be obtained, so we compare qualitatively to the results extracted from their paper (Figure~\ref{fig:comparison}, third column).

\iccvNew{Table~\ref{tab:metric_results} shows that our model consistently provides results closer to the ground truth scanpaths than Zhu et al.'s and SaltiNet. The latter is based on a saliency-sampling strategy, and thus these results indicate that indeed the temporal information learnt by our model is relevant for the final result.} \iccvNew{Our model, as expected, also amply surpasses the random baseline.} 
In Figure~\ref{fig:comparison} we see how PathGAN scanpaths fail to focus on the relevant parts of the scene (see, \eg, \emph{snow} or \emph{square}), while SaltiNet exhibits a somewhat erratic behavior, with large displacements and scarce areas of focus (\emph{train}, \emph{snow} or \emph{square} show this).
Finally, Zhu et al.'s approach tends to place gaze points at high contrast borders (see, \eg, \emph{square} or \emph{resort}).

\begin{figure}[t]
	\centering
	\includegraphics[width=\linewidth]{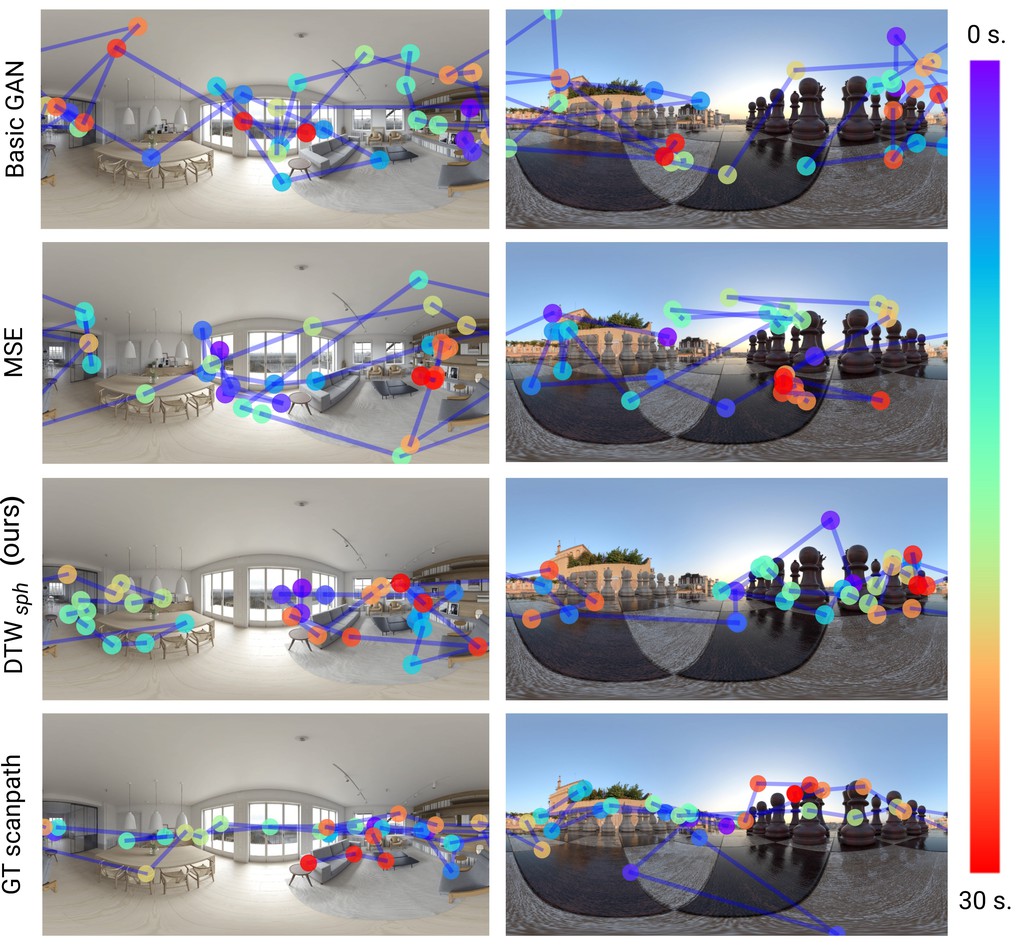}
	\caption{\iccvNew{Qualitative ablation results.} 
	\iccvOut{for two images from the Sitzmann et al.'s test set. }
	\emph{From top to bottom}: basic GAN strategy \iccvNew{(baseline)}; adding MSE to the loss function of the former; our approach; and an example ground truth scanpath. These results illustrate the need for our DTW$_{sph}$ loss term. 
	}
	\label{fig:ablation}
\end{figure}

\subsection{Ablation Studies}
\label{subsec:model_ablation}

We also evaluate the contribution of different elements of our model to the final result. For this purpose, we analyze a standard 
GAN strategy (\ie, using only the discriminative loss), as the baseline. Figure~\ref{fig:ablation} shows how the model is unable to learn both the temporal nature of the scanpaths, and their relation to image features. 
We also analyze the results yielded by adding a term based on the MSE between the ground truth and the generated scanpath to the loss function, instead of our DTW$_{sph}$ term (the only previous GAN approach for scanpath generation~\cite{assens2018pathgan} relied on MSE for their loss term). The MSE only measures a one-to-one correspondence between points, considering for each time instant a single point, unrelated to the rest. This hinders the learning process, leading to non-plausible results (Figure~\ref{fig:ablation}, second row). 
This behavior is corrected when our DTW$_{sph}$ is added instead, since it is specifically targeted for time series data and takes into account the actual spatial structure of the data (Figure~\ref{fig:ablation}, third row). The corresponding quantitative measures over our 
test set from Sitzmann et al. can be found in Table~\ref{tab:ablation}. We also analyze the effect of removing the CoordConv layer from our model: Results in Table~\ref{tab:ablation} indicate that the use of CoordConv does have a positive effect on the results, helping learn the transformation from the input to the target domain.

\begin{table}[t!]
\centering
\caption{\iccvNew{Quantitative comparisons of our model against %
SaltiNet~\cite{assens2018scanpath} and Zhu et al.~\cite{zhu2018prediction}. We also include upper (human baseline, \emph{Human BL}) and lower (randomly sampling over the image, \emph{Random BL}) baselines. Arrows indicate whether higher or lower is better, and boldface highlights the best result for each metric (excluding the ground truth \emph{Human BL}). $^*$SaltiNet is trained with Rai et al.'s dataset; we include it for completeness.}}

\vspace{1em}
\label{tab:metric_results}
\arrayrulecolor[rgb]{0.753,0.753,0.753}
\resizebox{\columnwidth}{!}{%
\begin{tabular}{c!{\color{black}\vrule}c!{\color{black}\vrule}cccc}
Dataset                                                                                         & Method     & LEV $\downarrow$  & DTW $\downarrow$  & REC $\uparrow$  & DET $\uparrow$   \\ 
\arrayrulecolor{black}\hline
\multirow{3}{*}{\begin{tabular}[c]{@{}c@{}} \\ Test set from\\ Sitzmann et al. \end{tabular}} & Random BL   & 52.33                   & 2370.56          & 0.47            & 0.93           \\

\arrayrulecolor[rgb]{0.753,0.753,0.753}\cline{2-6}

& SaltiNet   & 48.00                        & 1928.85           & 1.45            & 1.78             \\
                                                                                                & ScanGAN360 (ours)       & \textbf{46.15}      & \textbf{1921.95}  & \textbf{4.82}   & \textbf{2.32}    \\ 
\arrayrulecolor[rgb]{0.753,0.753,0.753}\cline{2-6}
                                                                                                & Human BL         & 43.11                         & 1843.72           & 7.81            & 4.07             \\ 
\arrayrulecolor{black}\hline
\multirow{3}{*}{\begin{tabular}[c]{@{}c@{}} \\ Rai et al.'s\\ dataset \end{tabular}}     

& Random BL   & 43.11                   & 1659.75          & 0.21            & 0.94           \\

\arrayrulecolor[rgb]{0.753,0.753,0.753}\cline{2-6}

& SaltiNet$^{*}$ & 48.07                & 1928.41           & 1.43            & 1.81             \\
& Zhu et al. & 43.55                & 1744.20           & 1.64            & 1.50             \\
                                                                                                & ScanGAN360 (ours)       & \textbf{40.99}            & \textbf{1549.59}  & \textbf{1.72}   & \textbf{1.87}    \\ 
\arrayrulecolor[rgb]{0.753,0.753,0.753}\cline{2-6}
                                                                                                & Human BL         & 39.59                        & 1495.55           & 2.33            & 2.31            
\end{tabular}
}
\arrayrulecolor{black}
\vspace{-0.6em}
\end{table}

\begin{table}[t]
\centering
\caption{\iccvNew{Quantitative results of our ablation study.
Arrows indicate whether higher or lower is better, and boldface highlights the best result for each metric (excluding the ground truth \emph{Human BL}). Please refer to the text for details on the ablated models.} 
\iccvOut{We evaluate the following: a basic GAN strategy (Basic GAN), the addition of a MSE error instead of the DTW$_{sph}$ term (MSE), and the effect of using CoordConv in our model. Qualitative results of this ablation study can be seen in Figure~\ref{fig:ablation}. Best results among the tested models are highlighted in bold for each metric.}
}
\vspace{1em}
\label{tab:ablation}
\arrayrulecolor{black}
\resizebox{\columnwidth}{!}{%
\begin{tabular}{c|cccc}
Metric                 & LEV $\downarrow$   & DTW $\downarrow$  & REC $\uparrow$  & DET $\uparrow$   \\ 
\hline
Basic GAN              & 49.42                       & 2088.44           & 3.01            & 1.74             \\
MSE                  & 48.90             & 1953.21           & 2.41            & 1.73             \\
DTW$_{sph}$ (no CoordConv) & 47.82                 & 1988.38           & 3.67            & 1.99             \\ 
\arrayrulecolor[rgb]{0.753,0.753,0.753}\hline
DTW$_{sph}$ ({ours})                & \textbf{46.19 }        & \textbf{1925.20 } & \textbf{4.50 }  & \textbf{2.33 }   \\ 
\arrayrulecolor{black}\hline
Human Baseline ({Human BL})         & 43.11               & 1843.72           & 7.81            & 4.07            
\end{tabular}
}
\arrayrulecolor{black}
\end{table}

\subsection{Behavioral Evaluation}
\label{subsec:new_behavioral}

\begin{figure*}[t!]
	\centering
	\includegraphics[width=\linewidth]{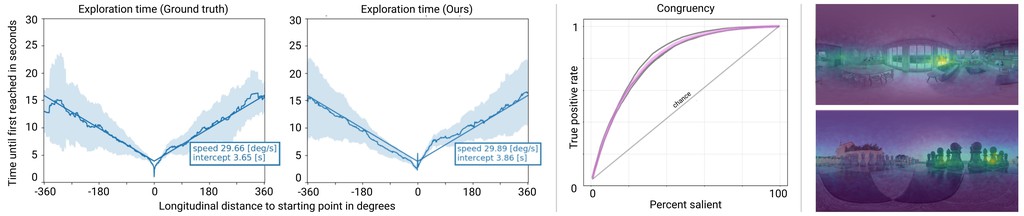}
	\caption{Behavioral evaluation. \emph{Left}: Exploration time 
	for real captured data (\emph{left}) and scanpaths generated by our model (\emph{center left}). Speed and exploration time of our scanpaths are on par with that of real users. \emph{Center right}: ROC curve of our generated scanpaths %
	for each individual test scene (gray), and averaged across scenes (magenta). The faster it converges to the maximum rate, the higher the inter-observer congruency. \emph{Right}: Aggregate maps for two different scenes, computed as heatmaps from 1,000 generated scanpaths. Our model is able to produce aggregate maps that focus on relevant areas of the scenes and exhibit the equator bias reported in the literature.}
	\label{fig:exploTime}
\end{figure*}

\iccvNew{While the previous subsections employ well-known metrics from the literature to analyze the performance of our model, in this subsection we perform a higher-level analysis of its results. We assess whether the behavioral characteristics of our scanpaths match those which have been reported from actual users watching 360$^\circ$ images.}

\paragraph{Exploration time} 
Sitzmann et al.~\cite{sitzmann2018saliency} measure the \emph{exploration time} as the average time that users took to move their eyes to a certain longitude relative to their starting point, and measure how long it takes for users to fully explore the scene. Figure~\ref{fig:exploTime} (left) shows this exploration time, measured by Sitzmann et al. from captured data, for the three scenes from their dataset included in our test set (\emph{room}, \emph{chess}, and \emph{robots}). To analyze whether our generated scanpaths mimic this behavior and exploration speed, we plot the exploration time of our generated scanpaths (Figure~\ref{fig:exploTime}, center left) for the same scenes and number of scanpaths. We can see how the speed and exploration time are very similar between real and generated data. Individual results per scene can be found in the supplementary material.

\paragraph{Fixation bias} 
\iccvNew{Similar to the center bias of human eye fixations observed in regular images~\cite{judd2009learning}, the existence of a Laplacian-like equator bias has been measured in 360$^\circ$ images~\cite{sitzmann2018saliency}:} The majority of fixations fall around the equator, in detriment of the poles. 
We have evaluated whether the distribution of scanpaths generated by our model also presents this bias. This is to be expected, since the data our model is trained with exhibits it, but is yet another indicator that we have succeeded in learning the ground truth distribution. We test this by generating, for each scene, 1,000 different scanpaths with our model, and aggregating them over time to produce a \emph{pseudo-}saliency map, which we term \emph{aggregate map}. Figure~\ref{fig:exploTime} (right) 
shows this for two scenes in our test set: We can see how this equator bias is indeed present in our generated scanpaths.

\paragraph{Inter-observer congruency} It is common in the literature analyzing users' gaze behavior to measure inter-observer congruency, often by means of a receiver operating characteristic (ROC) curve. We compute the congruency of our ``generated observers'' through this ROC curve for the three scenes in our test set from the Sitzmann et al. dataset (Figure~\ref{fig:exploTime}, center right). The curve calculates the ability of the $i^{th}$ scanpath to predict the \emph{aggregate map} of the corresponding scene.  
Each point in the curve is computed by generating a map containing the top $n\%$ most salient regions of the aggregate map (computed without the $i^{th}$ scanpath), and calculating the percentage of gaze points of the $i^{th}$ scanpath that fall into that map. Our ROC curve indicates a strong agreement between our scanpaths, with around $75\%$ of all gaze points falling within $25\%$ of the most salient regions. These values are comparable to those measured in previous studies with captured gaze data~\cite{sitzmann2018saliency,lemeur2013}.

\paragraph{Temporal and spatial coherence} 
Our generated scanpaths have a degree of stochasticity, to be able to model the diversity of real human observers. However, human gaze behavior follows specific patterns, and each gaze point is conditioned not only by the features in the scene but also by the previous history of gaze points of the user. If two users start watching a scene in the same region, a certain degree of coherence between their scanpaths is expected, that may diverge more as more time passes. We analyze the temporal coherence of generated scanpaths that start in the same region\iccvNew{, and observe that indeed our generated scanpaths follow a coherent pattern.
Please refer to the supplementary for more information on this part of the analysis.} 

\iccvOut{Specifically, we do this by generating a set of random scanpaths for each of the scenes in our test dataset, and separate them according to the longitudinal region where the scanpath begins (\eg, $[0^\circ, 40^\circ), [40^\circ,80^\circ)$, etc.). Then, we estimate the probability density of the generated scanpaths from each starting region using kernel density estimation (KDE) for each timestamp. \dani{This is NO LONGER shown in Figure~\ref{fig:kde} for two different starting regions of the $chess$ scene, at four different timestamps, and computed over 1000 generated scanpaths}. During the first seconds (first column), gaze points tend to stay in a smaller area, and closer to the starting region; as time progresses, they exhibit a more exploratory behavior with higher divergence, and eventually may reach a convergence close to regions of interest. We can also see how the behavior can differ depending on the starting region: at 10 seconds (second column), in the top row most of the attention is focused on the black pieces, while in the bottom row the attention is split between two large regions. The supplementary video and Section S8 in the supplementary material contain more results, including different scenes and the whole set of starting regions. }

\section{Conclusion}
\label{sec:conclusion}

In summary, we propose \iccvNew{ScanGAN360,} a conditional GAN approach to generating gaze scanpaths for immersive virtual environments. Our unique parameterization tailored to panoramic content, coupled with our novel usage of a DTW loss function, allow our model to generate scanpaths of significantly higher quality and duration than previous approaches. \iccvNew{We further explore applications of our model: Please refer to the supplementary material for a description and examples of these.}

Our GAN approach is well suited for the problem of scanpath generation: A \emph{single} ground truth scanpath does not exist, yet real scanpaths follow certain patterns that are difficult to model explicitly but that are automatically learned by our approach. Note that our model is also very fast and can produce about 1,000 scanpaths per second. This may be a crucial capability for interactive applications\iccvNew{: our model can generate \emph{virtual observers} in real time.}

\paragraph{Limitations and future work} 
\iccvOut{We train our model on Sitzmann et al.'s dataset~\cite{sitzmann2018saliency}. There is currently a lack of datasets of 360$^\circ$ images containing raw gaze positions. However, data-driven models strongly depend on the amount, diversity, and quality of the data they are trained with. Thus, more and better training data will likely help further improve the quality and generalization capabilities of our model. }
Our model is trained with 30-second long scanpaths, sampled at 1~Hz. Although this is significantly longer than most previous approaches
~\cite{ellis1985patterns, lemeur2013, liu2013semantically}, exploring different or variable lengths or sampling rates remains interesting for future work. 
When training our model, we focus on learning higher-level aspects of visual behavior, and we do not explicitly enforce low-level ocular movements (\eg,  fixations or saccades). 
Currently, our relatively low sampling rate prevents us from modeling very fast dynamic phenomena, such as saccades. Yet, fixation patterns naturally emerge in our results, and future work could explicitly take low-level oculomotor aspects of visual search into account.

\iccvOut{We have described proof-of-concept applications for our model, for example by showing that our scanpaths adapt their behavior to the configuration of the scene in Section~\ref{sec:applications}. Further development and optimization of these applications remains an interesting avenue for future work.} 

The model, parameterization, and loss function are tailored to 360$^\circ$ images. In a similar spirit, a DTW-based loss function could also be applied to conventional 2D images (using an Euclidean distance in 2D instead of our $\delta_{sph}$), potentially leading to better results than current 2D approaches based on mean-squared error.

We believe that our work is a timely effort and a first step towards understanding and modeling dynamic aspects of attention in 360$^{\circ}$ images. We hope that our work will serve as a basis to advance this research, both in virtual reality and in conventional imagery, and extend it to
other scenarios, such as dynamic or interactive content, analyzing the influence of the task, including the presence of motion parallax, or exploring multimodal experiences. We will make our model and training code available in order to facilitate the exploration of these and other possibilities.

{\small
\bibliographystyle{ieee_fullname}
\bibliography{references}
}

\clearpage

\begin{titlepage}
\maketitle
\end{titlepage}

\renewcommand\thesection{}
\section*{Supplementary Material}

\iccvNew{This document offers additional information and details on the following topics:}
\begin{itemize}
    \item (S1) Extended description of the soft-DTW (differentiable version of DTW) distance metric used in our model.
	\item (S2) Additional results (scanpaths generated with our method) for different scenes used in our evaluation in the main paper.
	\item (S3) Additional ground truth scanpaths for the scenes used in our evaluation in the main paper.
	\item (S4) Further details on our training process.
	\item (S5) Further details on metrics and evaluation, including a larger set of metrics (which we briefly introduce), and extended analysis.
	\item (S6) Further details on the behavioral evaluation of our scanpaths.
	\item (S7) Example applications of our method.
\end{itemize}

\setcounter{section}{0}
\renewcommand\thesection{S\arabic{section}}

\section{Differentiable Dynamic Time Warping: soft-DTW}
\label{sec:supp_softDTW}

\begin{figure}
    \centering
    \includegraphics[width=\linewidth]{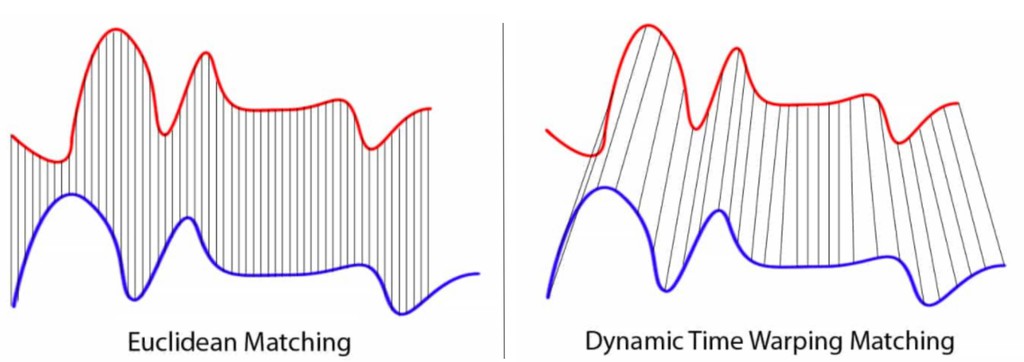}
    \caption{Simple visualization of dynamic time warping (DTW) alignment. Instead of assuming a pair-wise strict correspondence, DTW optimizes the alignment between two sequences to minimize their distance.}
    \label{fig:supp_dtw}
\end{figure}

One of the key aspects of our framework relies in the addition of a second term to the generator's loss function, based on dynamic time warping~\cite{muller2007dynamic}. As pointed in Section 3.3 in the main paper, dynamic time warping (DTW) measures the similarity between two temporal sequences (see Figure~\ref{fig:supp_dtw}\footnote{https://databricks.com/blog/2019/04/30/understanding-dynamic-time-warping.html}, Equation 5 in the main paper for the original DTW formulation, and Equations 6 and 7 in the main paper for our spherical modification on DTW). However, the original DTW function is not differentiable, therefore it is not suitable as a loss function. Instead, we use a differentiable version of it, soft-DTW, which has been recently proposed~\cite{cuturi2017soft} and used as a loss function in different problems dealing with time series~\cite{blondel2020differentiable, Chang_2019_CVPR, vincent2019shape}. 

Differently from the original DTW formulation (Equation 5 in the main paper), the soft-DTW is defined as follows:

\begin{equation}
    \textup{soft-DTW$^{\gamma}$}(\mathbf{r}, \mathbf{s}) = {\min_A}^{\gamma} \langle A, \Delta(\mathbf{r}, \mathbf{s}) \rangle,
    \label{eq:supp_soft-dtw}
\end{equation}

where, as with traditional DTW, $\Delta(\mathbf{r}, \mathbf{s}) = [\delta(r_i,s_j]_{ij} \in \mathbb{R}^{n \times m}$ is a matrix containing the distances $\delta(\cdot,\cdot)$ between each pair of points in $\mathbf{r}$ and $\mathbf{s}$, $A$ is a binary matrix that accounts for the alignment (or correspondence) between $\mathbf{r}$ and $\mathbf{s}$, and $\langle\cdot,\cdot\rangle$ is the inner product between both matrices. In our case, $\mathbf{r}=(r_1,...,r_T)\in \mathbb{R}^{3 \times T}$ and $\mathbf{s}=(s_1,...,s_T)\in \mathbb{R}^{3 \times T}$
are two scanpaths that we wish to compare.

The main difference lies in the replacement of the  ${\min_A}$ with the ${\min_A}^{\gamma}$ function, which is defined as follows:

\begin{equation}
    \label{dtw_loss_min}
    {\min}^{\gamma}(a_1,...,a_N) = \left\{\begin{matrix}
    {\min}_{i\le n}\,a_i, &\gamma = 0 \\ 
    -\gamma \log \sum_{n}^{i=1} e^{-a_i/\gamma}, &\gamma > 0
    \end{matrix}\right.
\end{equation}

This soft-min function allows DTW to be differentiable, with the $\gamma$ parameter adjusting the similarity between the soft implementation and the original DTW algorithm, both being the same when $\gamma = 0$.

\section{Additional Results}
\label{sec:supp_times}

We include in this section a more extended set of results. First, we include results for the scenes \textit{room} (see Figures~\ref{fig:ex1_1} to~\ref{fig:ex1_4}), \textit{chess} (see Figures~\ref{fig:ex2_1} to~\ref{fig:ex2_4}), and \textit{robots} (see Figures~\ref{fig:ex3_1} to~\ref{fig:ex3_4}) from the Sitzmann et al. dataset~\cite{sitzmann2018saliency}. Then, we include results for the five scenes from the Rai et al. dataset~\cite{rai2017dataset} used in comparisons throughout the main paper: \textit{train} (see Figures~\ref{fig:ex4_1} to~\ref{fig:ex4_4}), \textit{resort} (see Figures~\ref{fig:ex5_1} to~\ref{fig:ex5_4}), \textit{square} (see Figures~\ref{fig:ex6_1} to~\ref{fig:ex6_4}), \textit{snow} (see Figures~\ref{fig:ex7_1} to~\ref{fig:ex7_4}), and \textit{museum} (see Figures~\ref{fig:ex8_1} to~\ref{fig:ex8_4}).

\section{Ground Truth Scanpaths for Comparison Scenes}
\label{sec:supp_times}

We include in Figures~\ref{fig:supp_gt_comp_1} to \ref{fig:supp_gt_comp_5} sets of ground truth scanpaths for all the images shown in Figure 4 in the main paper, which is devoted to comparisons of our method against other models; and in Figures~\ref{fig:gt_sitz_1} to \ref{fig:gt_sitz_3} sets of ground truth scanpaths for the three images from our test set from Sitzmann et al.'s dataset.

\section{Additional Details on our Training Process}
\label{sec:supp_add_details}

In addition to the details commented in Section 3.5 in the main paper, our generator trains two cycles per discriminator cycle, to avoid the latter from surpassing the former. To enhance the training process, we also resort to a mini-batching strategy: Instead of inputting to our model a set containing all available scanpaths for a given image, we split our data in different mini-batches of eight scanpaths each. This way, the same image is input in our network multiple times per epoch, also allowing more images to be included in the same batch, and therefore enhancing the training process. We trained our model for 217 epochs, as we found that epoch to yield the better evaluation results.

\section{Additional Details on Metrics and Evaluation}
\label{sec:supp_metrics}

Throughout this work, we evaluate our model and compare to state-of-the-art works by means of several widely used metrics, recently reviewed by Fahimi and Bruce~\cite{fahimi2020metrics}.
Table~\ref{tab:supp_metrics} shows a list of these metrics, indicating which ones take into account position and/or order of gaze points. \iccvNew{In the following, we briefly introduce these metrics (please refer to Fahimi and Bruce~\cite{fahimi2020metrics} for a formal description):}  

\begin{itemize}
	\item Levenshtein distance: Transforms scanpaths into strings, and then calculates the minimum number of single-character edits (insertions, deletions, or substitutions) required to change one \textit{string} (scanpath) into the other. All edits costs are treated equally.
	\item ScanMatch: Improved version of Levenshtein distance. Different from Levenshtein distance, ScanMatch takes into account semantic information (as a score matrix), and can even take into account duration of data points. This way, each of the edit operations can be differently weighted. 
	\item Hausdorff distance: Represents the degree of mismatch between two sets by measuring the farthest spatial distance from one set to the other, i.e., the distance between two different curves.
	\item Frechet distance: Similar to Hausdorff distance, it measures the similarity between curves. However, Frechet disatance takes into account both the position and ordering of all the points in the curves.
	\item Dynamic time warping: Metric that compares two
time-series with varying (and differing) lengths to find an optimal path to match both sequences while preserving boundary, continuity, and monotonicity to make sure that the path respects time.
	\item Time delay embedding: Splits a scanpath into several sub-samples, i.e., small sub-scanpaths. This metrics calculates a similarity score by performing several pair-wise Hausdorff comparisons over sub-samples from both scanpaths to compare.
	\item Recurrence: Measures the percentage of gaze points that match (are close) between the two scanpaths.
	\item Determinism: Percentage of cross-recurrent points that form diagonal lines (i.e., percentage of gaze trajectories common to both scanpaths).
	\item Laminarity: Measures locations that were fixated in detail in one of the scanpaths, but only fixated briefly in the other scanpath. This way, it indicates whether specific areas of a scene are repeatedly fixated.
	\item Center of recurrence mass: Defined as the distance of the center of gravity from the main diagonal, indicates the dominant lag of cross recurrences, i.e., whether the same gaze point in both scanpaths tends to occur close in time.
\end{itemize}

\begin{table}[ht]
\caption{Set of metrics to quantitatively evaluate scanpath similarity~\cite{fahimi2020metrics}. Each metric specializes in specific aspects of the scanpaths, and as a result using any of them in isolation may not be representative.}
\vspace{1em}
\begin{tabular}{lccc}
\hline
\multicolumn{1}{c}{\textbf{Metric}} & \textbf{Abrv} & \textbf{Position}                     & \textbf{Order}                        \\ \hline
Levenshtein distance		        & LEV           & \textbf{\color{ticgreen}{\cmark}} & \textbf{\color{ticgreen}{\cmark}} \\
ScanMatch                           & SMT           & \textbf{\color{ticgreen}{\cmark}} & \textbf{\color{ticgreen}{\cmark}} \\
Hausdorff distance                  & HAU           & \textbf{\color{ticgreen}{\cmark}} & \textit{\color{red}{\xmark}}                 \\
Frechet distance                    & FRE           & \textbf{\color{ticgreen}{\cmark}} & \textbf{\color{ticgreen}{\cmark}} \\
Dynamic time warping \, \, \, \, \, & DTW           & \textbf{\color{ticgreen}{\cmark}} & \textbf{\color{ticgreen}{\cmark}} \\
Time delay embedding                & TDE           & \textbf{\color{ticgreen}{\cmark}} & \textit{\color{red}{\xmark}}                 \\
Recurrence                          & REC           & \textbf{\color{ticgreen}{\cmark}} & \textit{\color{red}{\xmark}}                 \\
Determinism                         & DET           & \textit{\color{red}{\xmark}}                 & \textbf{\color{ticgreen}{\cmark}} \\
Laminarity                          & LAM           & \textit{\color{red}{\xmark}}                 & \textit{\color{red}{\xmark}}                 \\
Center of recurrence mass          & COR           & \textit{\color{red}{\xmark}}                 & \textit{\color{red}{\xmark}}                 \\ \hline
\end{tabular}
\label{tab:supp_metrics}
\end{table}

Our model is stochastic by nature. This means that the scanpaths that it generates for a given scene are always different, simulating observer variability. We have analyzed whether the reported metrics vary depending on the number of scanpaths generated, to asses the stability and overall goodness of our model. Results can be seen in Table~\ref{tab:supp_diff_met}

\begin{table}[t]
\caption{Quantitative results of our model with sets of generated scanpaths with different number of samples. Our results are stable regardless the number of generated samples.}
\label{tab:supp_diff_met}
\vspace{1em}
\resizebox{\columnwidth}{!}{%
\arrayrulecolor{black}
\begin{tabular}{c|c|cccc}
Dataset                                                                                        & \# of samples & LEV $\downarrow$  & DTW $\downarrow$  & REC $\uparrow$  & DET $\uparrow$   \\ 
\hline
\multirow{4}{*}{\begin{tabular}[c]{@{}c@{}}Test set from\\ Sitzmann et al. \end{tabular}} & 100           & 46.19             & 1925.20           & 4.50            & 2.33             \\
                                                                                               & 800           & 46.10             & 1916.26           & 4.75            & 2.34             \\
                                                                                               & 2500          & 46.15             & 1921.95           & 4.82            & 2.32             \\ 
\arrayrulecolor[rgb]{0.753,0.753,0.753}\cline{2-6}
                                                                                               & Human BL            & 43.11             & 1843.72           & 7.81            & 4.07             \\ 
\arrayrulecolor{black}\hline
\multirow{4}{*}{\begin{tabular}[c]{@{}c@{}}Rai et al.'s\\ dataset \end{tabular}}          & 100           & 40.95             & 1548.86           & 1.91            & 1.85             \\
                                                                                               & 800           & 40.94             & 1542.82           & 1.86            & 1.86             \\
                                                                                               & 2500          & 40.99             & 1549.59           & 1.72            & 1.87             \\ 
\arrayrulecolor[rgb]{0.753,0.753,0.753}\cline{2-6}
                                                                                               & Human BL            & 39.59             & 1495.55           & 2.33            & 2.31            
\end{tabular}
}
\arrayrulecolor{black}
\end{table}

We include in Table~\ref{tab:supp_all_metrics} the evaluation results with the full set of metrics shown in Table~\ref{tab:supp_metrics} (extension to Table 1 in the main paper), and in Tables~\ref{tab:supp_abl} and~\ref{tab:supp_abl2} the evaluation results of our ablation studies over the full set of metrics (extension to Table 2 in the main paper).

\begin{table*}[ht!]
\centering
\caption{Quantitative comparison of our model against different approaches, following the metrics introduced in Table 1. We evaluate our model over the test set we separated from Sitzmann et al.'s dataset, and compare against Saltinet~\cite{assens2017saltinet}. On the other hand, we validate our model over Rai et al.'s dataset, and compare us against Zhu et al.'s~\cite{zhu2018prediction}, whose results over this dataset were provided by the authors; and against SaltiNet, which was trained over that specific dataset (*). HB accounts for the human baseline, computed with the set of ground-truth scanpaths. \iccvNew{We also compute a lower baseline, computed by randomly sampling the image.}
The arrow next to each metric indicates whether higher or lower is better. Best results are in bold.}
\label{tab:supp_all_metrics}
\vspace{1em}
\arrayrulecolor[rgb]{0.753,0.753,0.753}
\resizebox{2\columnwidth}{!}{%
\begin{tabular}{c!{\color{black}\vrule}c!{\color{black}\vrule}cccccccccc}
Dataset                                                                                     & Method            & LEV $\downarrow$  & SMT $\uparrow$  & HAU $\downarrow$  & FRE $\downarrow$  & DTW $\downarrow$  & TDE $\downarrow$  & REC $\uparrow$  & DET $\uparrow$  & LAM $\uparrow$  & CORM $\uparrow$   \\ 
\arrayrulecolor{black}\hline
\multirow{4}{*}{\begin{tabular}[c]{@{}c@{}}\\Test set from\\ Sitzmann et al. \end{tabular}} & Random BL         & 52.33             & 0.22            & 59.88             & 146.39            & 2370.56           & 27.93             & 0.47            & 0.93            & 9.19            & 33.19             \\ 
\arrayrulecolor[rgb]{0.753,0.753,0.753}\cline{2-12}
                                                                                            & SaltiNet          & 48.00             & 0.18            & 64.23             & 149.34            & 1928.85           & 28.19             & 1.45            & 1.78            & 10.45           & 29.23             \\
                                                                                            & ScanGAN360 (ours) & \textbf{46.15}    & \textbf{0.39}   & \textbf{43.28}    & \textbf{141.23}   & \textbf{1921.95}  & \textbf{18.62}    & \textbf{4.82}   & \textbf{2.32}   & \textbf{24.51}  & \textbf{35.78}    \\ 
\cline{2-12}
                                                                                            & Human BL          & 43.11             & 0.43            & 41.38             & 142.91            & 1843.72           & 16.05             & 7.81            & 4.07            & 24.69           & 35.32             \\ 
\arrayrulecolor{black}\cline{2-12}
\multirow{5}{*}{\begin{tabular}[c]{@{}c@{}}Rai et al's\\ dataset \end{tabular}}             & Random BL         & 43.11             & 0.17            & 65.71             & 144.73            & 1659.75           & 35.41             & 0.21            & 0.94            & 4.30            & 19.08             \\ 
\arrayrulecolor[rgb]{0.753,0.753,0.753}\cline{2-12}
                                                                                            & SaltiNet$^{(*)}$  & 48.07             & 0.18            & 63.86             & 148.76            & 1928.41           & 28.42             & 1.43            & 1.81            & 10.22           & 29.33             \\
                                                                                            & Zhu et al.        & 43.55             & 0.20            & 73.09             & \textbf{136.37}   & 1744.20           & 30.62             & 1.64            & 1.50            & 9.18            & 26.05             \\
                                                                                            & ScanGAN360 (ours) & \textbf{40.99}    & \textbf{0.24}   & \textbf{61.86}    & 139.10            & \textbf{1549.59}  & \textbf{28.14}    & \textbf{1.72}   & \textbf{1.87}   & \textbf{12.23}  & \textbf{26.15}    \\ 
\cline{2-12}
                                                                                            & Human BL          & 39.59             & 0.24            & 66.23             & 136.70            & 1495.55           & 27.24             & 2.33            & 2.31            & 14.36           & 23.14            
\end{tabular}
}
\arrayrulecolor{black}
\end{table*}

\begin{table*}[ht!]
\centering
\caption{Results of our ablation study over Sitzmann et al.'s test set. We take a basic GAN strategy as baseline, and evaluate the effects of adding a second term into our generator's loss function. We ablate a model with an MSE error (as used in the only GAN approach for scanpath generation so far~\cite{assens2018pathgan}), and compare it against our spherical DTW approach. We also analyze the importance of the CoordConv layer, whhose absence slightly worsen the results. See Section 4 in the main paper for further discussion. Qualitative results of this ablation study can be seen in Figure 5 in the main paper.}
\label{tab:supp_abl}
\vspace{1em}
\arrayrulecolor{black}
\resizebox{2\columnwidth}{!}{%
\begin{tabular}{c|cccccccccc}
Metric                   & LEV $\downarrow$  & SMT $\uparrow$  & HAU $\downarrow$  & FRE $\downarrow$  & DTW $\downarrow$  & TDE $\downarrow$  & REC $\uparrow$  & DET $\uparrow$  & LAM $\uparrow$  & CORM $\uparrow$   \\ 
\hline
Basic GAN                           & 49.42             & 0.36            & 43.69             & 145.95            & 2088.44           & 20.05             & 3.01            & 1.74            & 18.55           & 34.51             \\
MSE                                & 48.90             & 0.37            & \textbf{42.27 }   & \textbf{133.24}            & 1953.21           & 19.48             & 2.41            & 1.73            & 18.47           & \textbf{37.34 }   \\
DTW$_{sph}$ (no CoordConv)               & 47.82             & 0.37            & 46.59             & 144.92            & 1988.38           & 20.13             & 3.67            & 1.99            & 18.09           & 35.66             \\ 
\arrayrulecolor[rgb]{0.753,0.753,0.753}\hline
DTW$_{sph}$ (ours)                   & \textbf{46.15 }   & \textbf{0.39 }  & 43.28             & 141.23            & \textbf{1921.95}           & \textbf{18.62 }   & \textbf{4.82 }  & \textbf{2.32}            & \textbf{24.21 } & 35.78             \\ 
\arrayrulecolor{black}\hline
Human BL                      & 43.11             & 0.43            & 41.38             & 142.91            & 1843.72           & 16.05             & 7.81            & 4.07            & 24.69           & 35.32            
\end{tabular}
}
\arrayrulecolor{black}
\end{table*}

\begin{table*}[ht!]
\centering
\caption{Results of our ablation study over Rai et al.'s dataset. We take a basic GAN strategy as baseline, and evaluate the effects of adding a second term into our generator's loss function. We ablate a model with an MSE error (as used in the only GAN approach for scanpath generation so far~\cite{assens2018pathgan}), and compare it against our spherical DTW approach. We also analyze the importance of the CoordConv layer, whhose absence slightly worsen the results. See Section 4 in the main paper for further discussion. Qualitative results of this ablation study can be seen in Figure 5 in the main paper.}
\label{tab:supp_abl2}
\vspace{1em}
\arrayrulecolor{black}
\resizebox{2\columnwidth}{!}{%
\begin{tabular}{c|cccccccccccc}
Metric                   & LEV $\downarrow$  & SMT $\uparrow$  & HAU $\downarrow$  & FRE $\downarrow$  & DTW $\downarrow$  & TDE $\downarrow$  & REC $\uparrow$  & DET $\uparrow$  & LAM $\uparrow$  & CORM $\uparrow$   \\ 
\hline
Basic GAN                            & 41.73             & 0.23           & 59.11           & 142.42           & 1542.52         & 28.40           & 0.99          & 1.47            & 8.08        & 24.55           \\ 

MSE                                & 41.81            & 0.23           & \textbf{61.30}            & 139.59           & \textbf{1541.44}         & 28.66            & 1.01          & 1.51           & 8.56        & 24.45           \\ \

DTW$_{sph}$ (no CoordConv)              & 41.42             & 0.23            & 61.55            & 148.13           & 1610.10        & 28.78           & 1.61          & 1.65            & 10.25        & 24.68           \\ 

\arrayrulecolor[rgb]{0.753,0.753,0.753}\hline

DTW$_{sph}$ (ours)                     & \textbf{40.99}    & \textbf{0.24}   & 61.86    & \textbf{139.10}   & 1549.59  & \textbf{28.14}    & \textbf{1.72}   & \textbf{1.87}   & \textbf{12.23}  & \textbf{26.15} \\ 

\arrayrulecolor{black}\hline
Human BL                      & 39.59             & 0.24            & 66.23             & 136.70            & 1495.55           & 27.24             & 2.33            & 2.31            & 14.36           & 23.14               
\end{tabular}
}
\arrayrulecolor{black}
\end{table*}

\iccvNew{Images for one of our test sets belong to Rai et al.'s dataset~\cite{rai2017dataset}. This dataset is larger than Sitzmann et al.'s in size (number of images), but provides gaze data in the form of fixations with associated timestamps, and not the raw gaze points. Note that most of the metrics proposed in the literature for scanpath similarity are designed to work with time series of different length, and do not necessarily assume a direct pairwise equivalence, making them valid to compare our generated scanpaths to the ground truth ones from Rai et al.'s dataset.}

\section{Behavioral Evaluation}
\label{sec:behavior}

\iccvNew{In this section, we include further analysis and additional details on behavioral aspects of our scanpaths, extending Section 4.5 in the main paper.}

\paragraph{Temporal and spatial coherence} 
As discussed in the main paper, our generated scanpaths have a degree of stochasticity, and different patterns arise depending on users' previous history. To assess whether our scanpaths actually follow a coherent pattern, we generate a set of random scanpaths for each of the scenes in our test dataset, and separate them according to the longitudinal region where the scanpath begins (\eg, $[0^\circ, 40^\circ), [40^\circ,80^\circ)$, etc.). Then, we estimate the probability density of the generated scanpaths from each starting region using kernel density estimation (KDE) for each timestamp. We include the complete KDE results for the three images from our test set in Figures~\ref{fig:supp_kde1}~to~\ref{fig:supp_kde6}, for different starting regions, at different timestamps, and computed over 1000 generated scanpaths. During the first seconds (first column), gaze points tend to stay in a smaller area, and closer to the starting region; as time progresses, they exhibit a more exploratory behavior with higher divergence, and eventually may reach a convergence close to regions of interest. We can also see how the behavior can differ depending on the starting region.

\begin{figure*}[t]
	\centering
	\includegraphics[width=\linewidth]{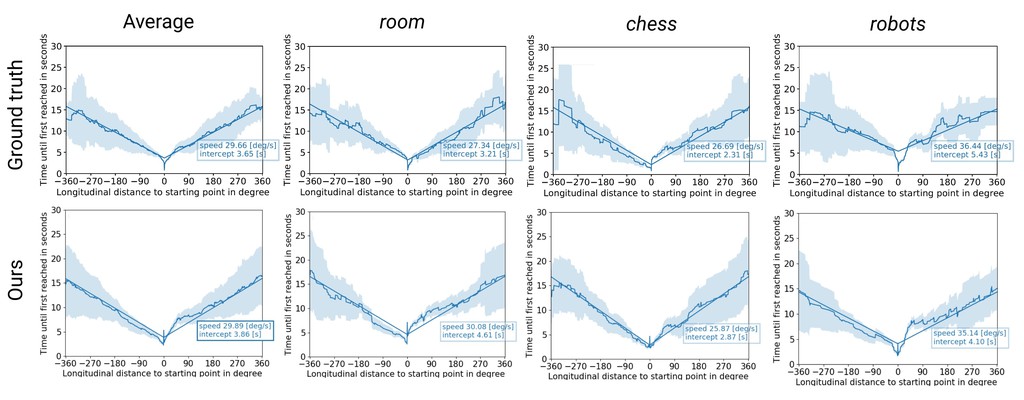}
	\caption{Time to explore each of the scenes from the Sitzmann et al. test set, together with their ground truth counterpart.}
	\label{fig:supp_time}
\end{figure*}

\paragraph{Exploration time} As introduced in the main paper, we also explore the time that users took to move their eyes to a certain longitude relative to their starting point, and measure how long it takes for users to fully explore the scene. We include in Figure~\ref{fig:supp_time} the comparison between ground truth and generated scanpaths in terms of time to explore the scene, for all the three scenes from our test set (\emph{room}, \emph{chess}, and \emph{robots}), both individual and aggregated. We can see how the speed and exploration time are very similar between real and generated data. 

\section{Applications of the Model}
\label{sec:applications}

Our model is able to generate plausible 30-second scanpaths, drawn from a distribution that mimics the behavior of human observers. \iccvNew{As we briefly discuss through the paper,} this enables a number of applications, starting with avoiding the need to recruit and measure gaze from high numbers of observers in certain scenarios. 
We show here two applications of our model, virtual scene design and scanpath-driven video thumbnail creation for static 360$^\circ$ images, and discuss other potential application scenarios.

\paragraph{Virtual scene design}

In an immersive environment, the user has control over the camera when exploring it. This poses a challenge to content creators and designers, who have to learn from experience how to layout the scene to elicit a specific viewing or exploration behavior. 
This is not only a problem in VR, but has also received attention in, \eg, manga composition~\cite{cao2014look} or web design~\cite{pang2016directing}.
However, actually measuring gaze from a high enough number of users to determine optimal layouts can be challenging and time-consuming. While certain goals may require real users, others can make use of our model to generate plausible and realistic generated observers.

\begin{figure*}[t]
	\centering
	\includegraphics[width=\linewidth]{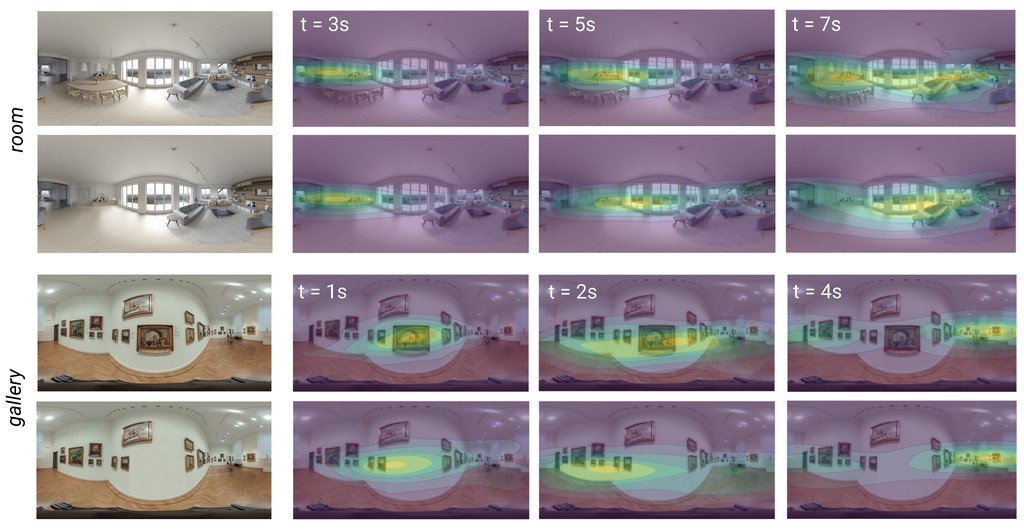}
	\caption{
    Our model can be used to aid the design of virtual scenes. We show two examples, each with two possible layouts (original, and removing some significant elements).  We generate a large number of scanpaths (virtual observers) starting from the same region, and compute their corresponding probability density function as a function of time, using KDE (see Section~\ref{sec:behavior}).
    \emph{room} scene: The presence of the dining table and lamps (top) retains the viewers' attention longer, while in their absence they move faster towards the living room area, performing a more linear exploration.             
    \emph{gallery} scene: When the central picture is present (top), the viewers linger there before splitting to both sides of the scene. In its absence, observers move towards the left, then explore the scene linearly in that direction.
	}
	\label{fig:att}
\end{figure*}

As a proof of concept, we have analyzed our model's ability to adapt its behavior to different layouts of a scene (Figure~\ref{fig:att}).
Specifically, we have removed certain elements from a scene, and run our model to analyze whether these changes affect the behavior of our generated scanpaths. We plot the resulting probability density (using KDE, see Section~\ref{sec:behavior}) as a function of time. The presence of different elements in the scene affects the general viewing behavior, including viewing direction, or time spent on a certain region. These examples %
are particularly promising if we consider that our model is trained with a relatively small number of generic scenes.

\paragraph{Scanpath-driven video thumbnails of static 360$^\circ$ images}

360$^\circ$ images capture the full sphere and are thus unintuitive when projected into a conventional 2D image. To address this problem, a number of approaches have proposed to retarget 360$^\circ$ images or videos to 2D~\cite{su2016pano2vid,sitzmann2018saliency,su2017making}. In the case of images, extracting a representative 2D visualization of the 360$^\circ$ image can be helpful to provide a thumbnail of it, for example as a preview on a social media platform. 
However, these thumbnails are static. The Ken Burns effect can be used to animate static images by panning and zooming a cropping window over a static image. In the context of 360$^\circ$, however, it seems unclear what the trajectory of such a moving window would be.

To address this question, we leverage our generated scanpaths to drive a Ken Burns--like video thumbnail of a static panorama. For this purpose, we use an average scanpath, computed as the probability density of several generated scanpaths using KDE (see Section~\ref{sec:behavior}), as the trajectory of the virtual camera. Specifically, KDE allows us to find the point of highest probability, along with its variance, of all generated scanpaths at any point in time. Note that this point is not necessarily the average of the scanpaths. We use the time-varying center point as the center of our 
2D viewport, 
and its variance to drive the FOV or zoom of the moving viewport. 

\begin{figure}[t]
	\centering
	\includegraphics[width=\linewidth]{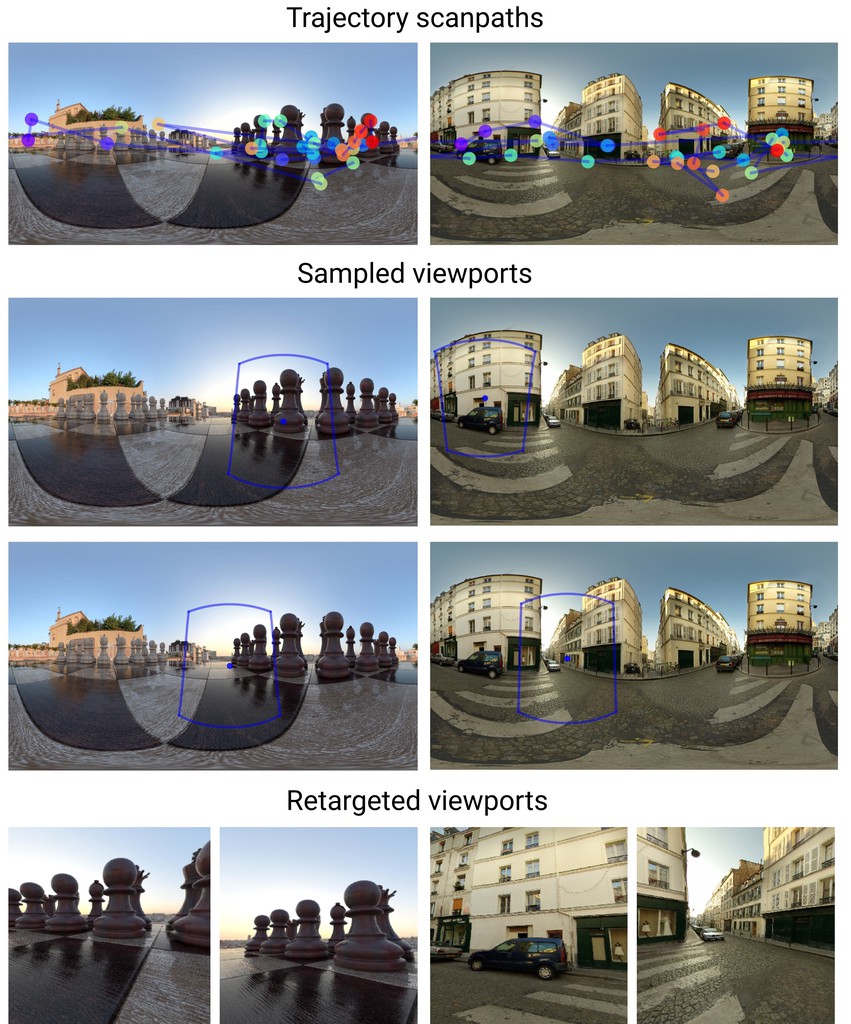}
	\caption{Scanpath-driven video thumbnails of 360$^\circ$ images. We propose a technique to generate these videos that results in relevant and intuitive explorations of the 360$^\circ$ scenes. Top row: Points of highest probability at each time instant, displayed as scanpaths. These are used as a guiding trajectory for the virtual camera. Middle rows: Two viewports from the guiding trajectory, corresponding to the temporal window with lowest variance.
	Bottom row: 2D images 
	retargeted from those viewports. Please refer to the text for details.
	}
	\label{fig:app_retargeting}
\end{figure}

\begin{figure*}[t]
	\centering
	\includegraphics[width=\linewidth]{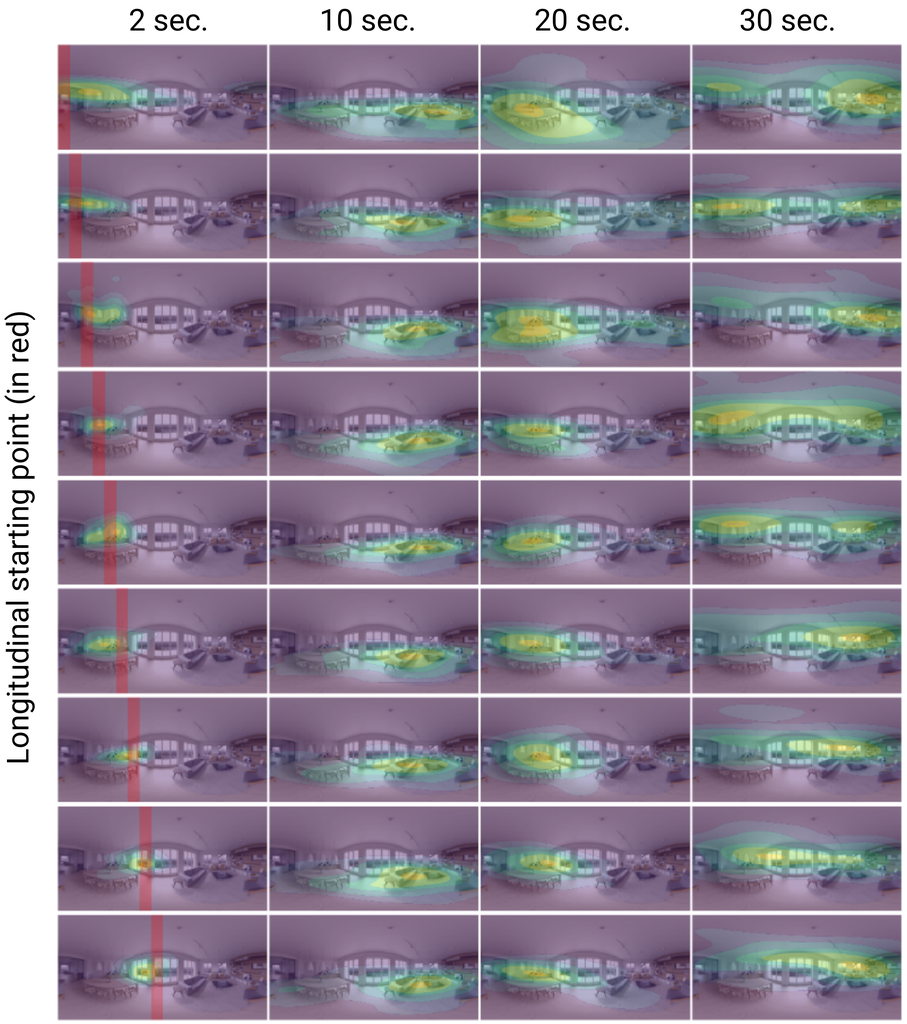}
	\caption{KDE for the \textit{room} scene, including scanpaths starting from $0^{\circ}$ up to $160^{\circ}$.  }
	\label{fig:supp_kde1}
\end{figure*}

\begin{figure*}[t]
	\centering
	\includegraphics[width=\linewidth]{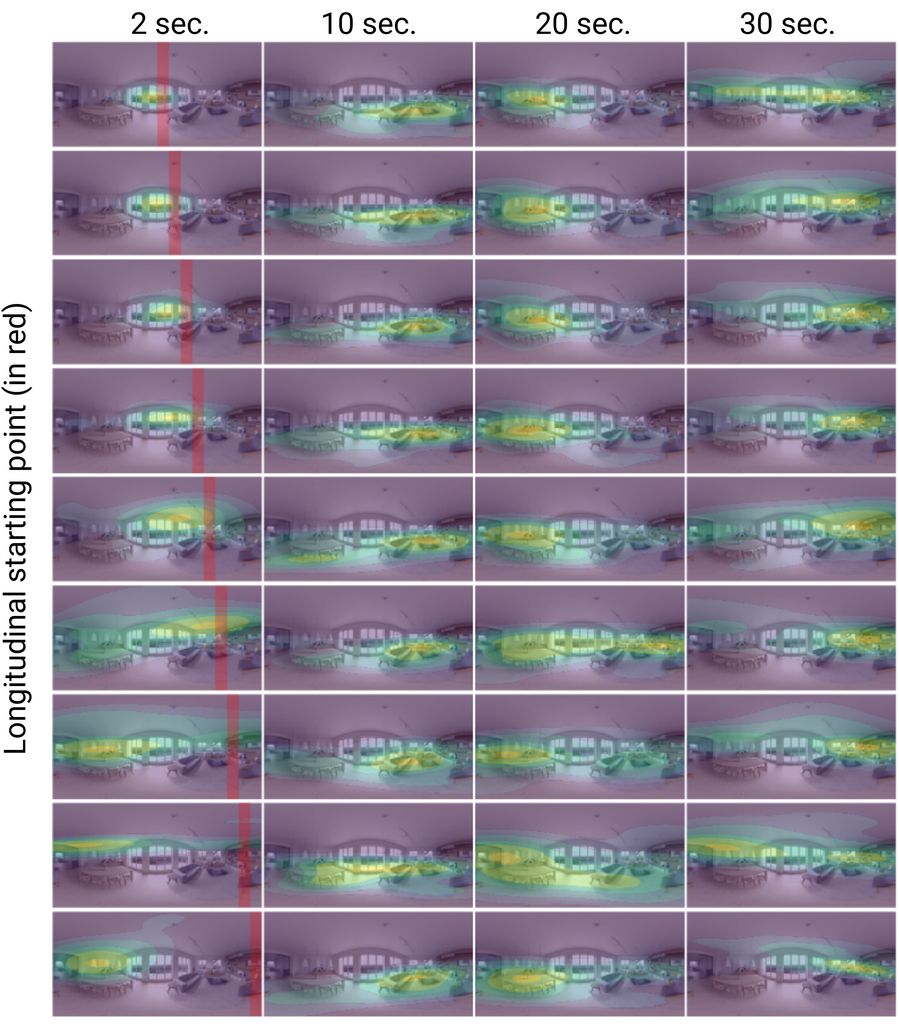}
	\caption{KDE for the \textit{room} scene, including scanpaths starting from $180^{\circ}$ up to $340^{\circ}$. }
	\label{fig:supp_kde2}
\end{figure*}

\begin{figure*}[t]
	\centering
	\includegraphics[width=\linewidth]{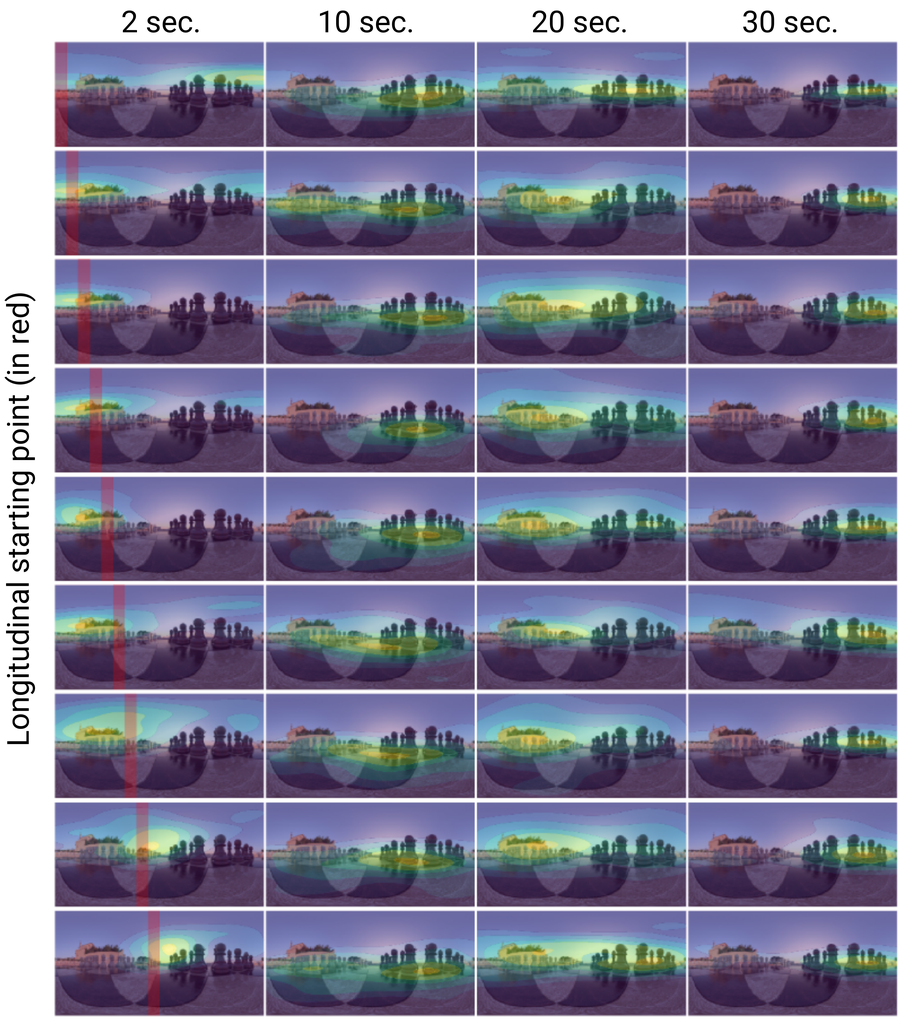}
	\caption{KDE for the \textit{chess} scene, including scanpaths starting from $0^{\circ}$ up to $160^{\circ}$. }
	\label{fig:supp_kde3}
\end{figure*}

\begin{figure*}[t]
	\centering
	\includegraphics[width=\linewidth]{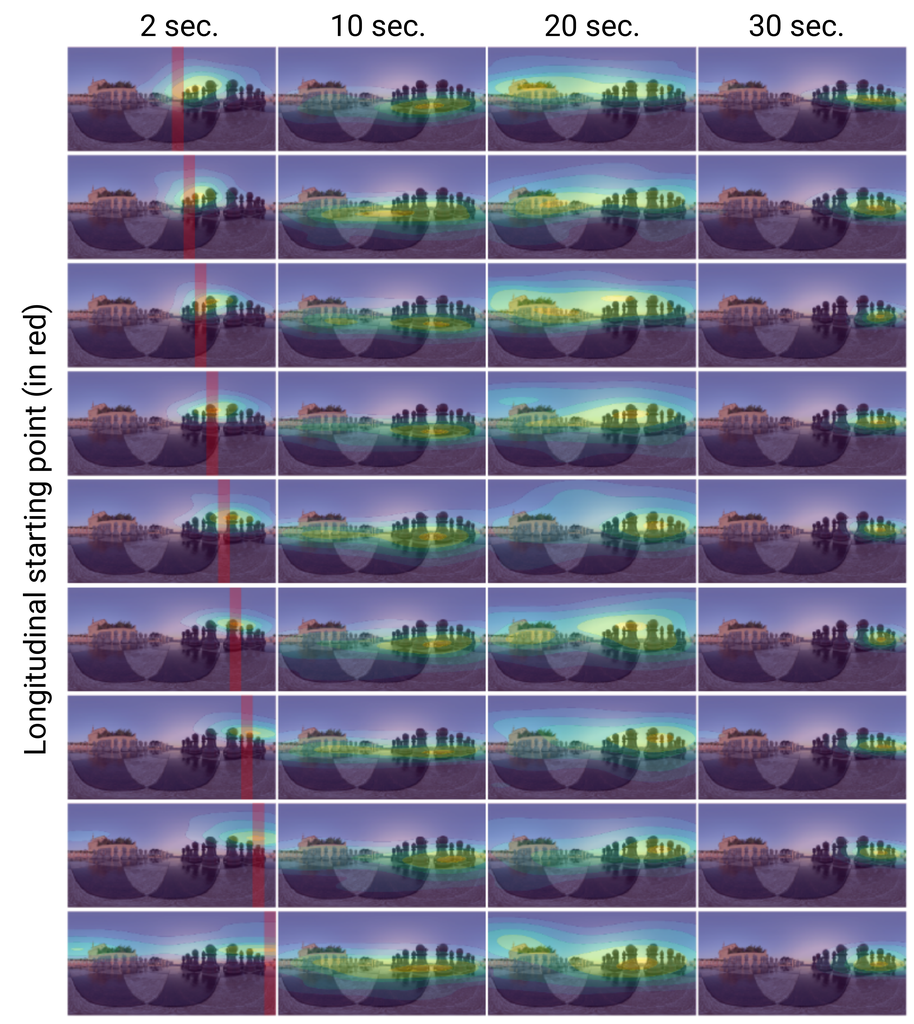}
	\caption{KDE for the \textit{chess} scene, including scanpaths starting from $180^{\circ}$ up to $340^{\circ}$. }
	\label{fig:supp_kde4}
\end{figure*}

\begin{figure*}[t]
	\centering
	\includegraphics[width=\linewidth]{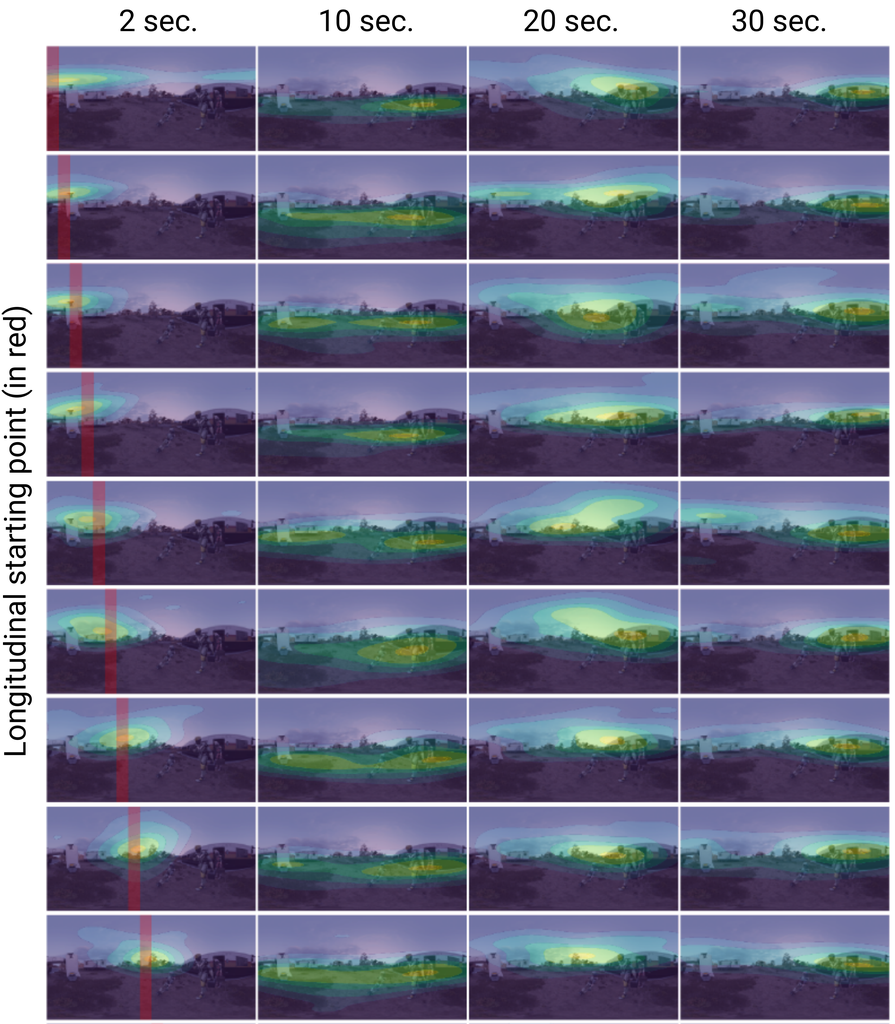}
	\caption{KDE for the \textit{robots} scene, including scanpaths starting from $0^{\circ}$ up to $160^{\circ}$. }
	\label{fig:supp_kde5}
\end{figure*}

\begin{figure*}[t]
	\centering
	\includegraphics[width=\linewidth]{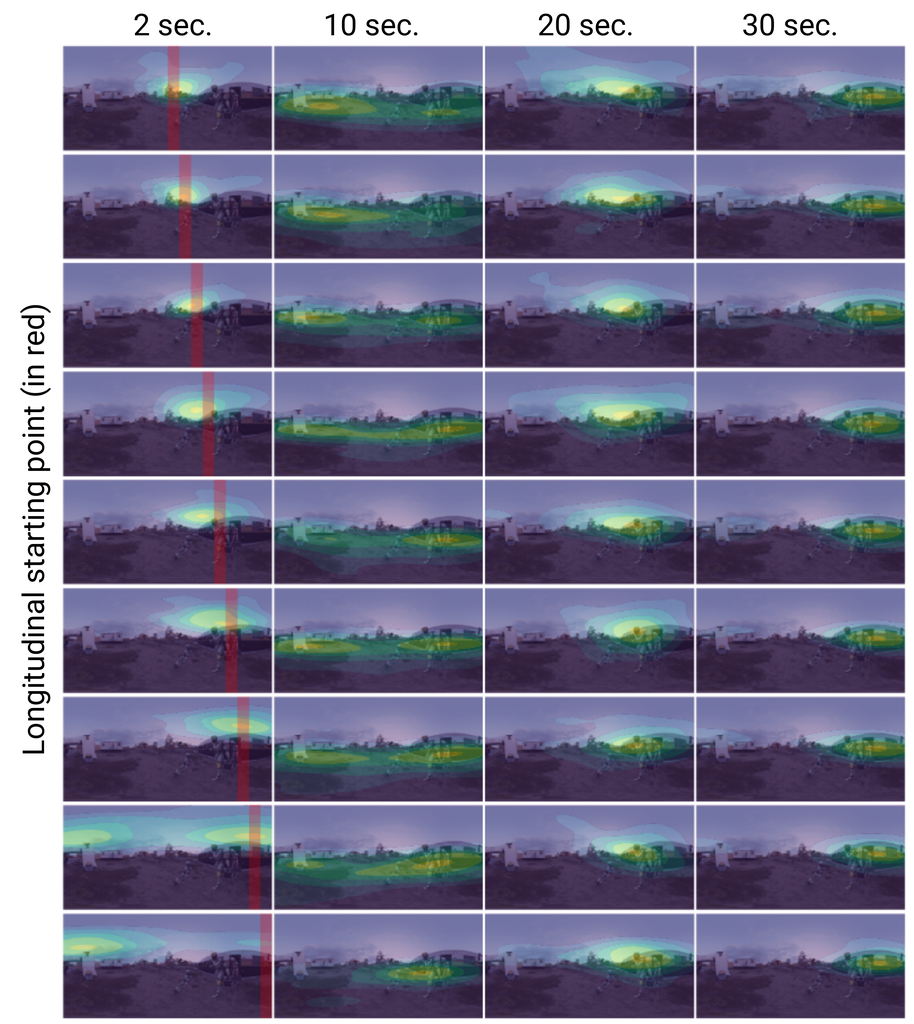}
	\caption{KDE for the \textit{robots} scene, including scanpaths starting from $180^{\circ}$ up to $340^{\circ}$. }
	\label{fig:supp_kde6}
\end{figure*}

\begin{figure*}[t]
	\centering
	\adjustbox{trim={.1\width} {.05\height} {0.1\width} {.05\height},clip,width=\linewidth}%
  {\includegraphics[width=\linewidth]{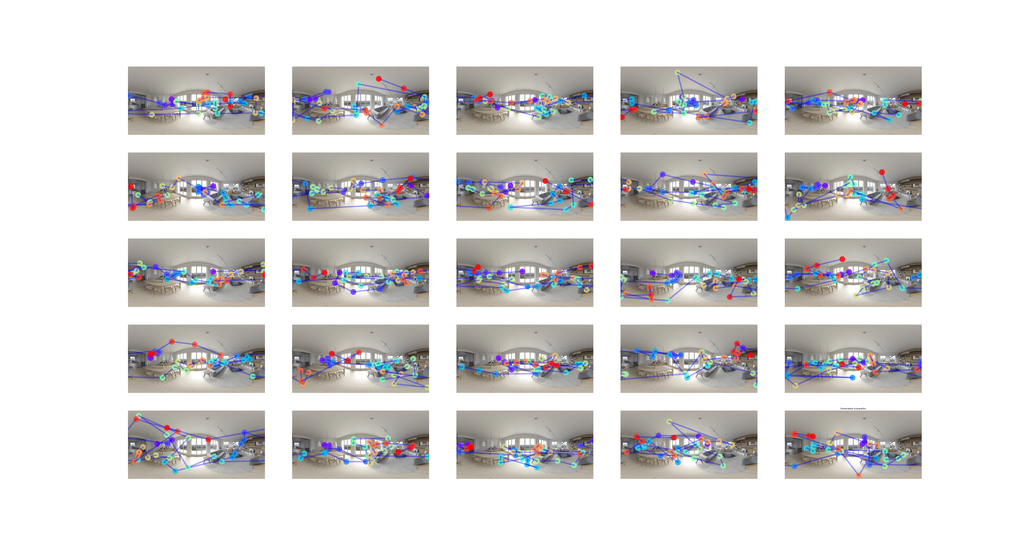}}
	\caption{Generated scanpaths for the \textit{room} scene.}
	\label{fig:ex1_1}
\end{figure*}

\begin{figure*}[t]
	\centering
	\adjustbox{trim={.1\width} {.05\height} {0.1\width} {.05\height},clip,width=\linewidth}%
  {\includegraphics[width=\linewidth]{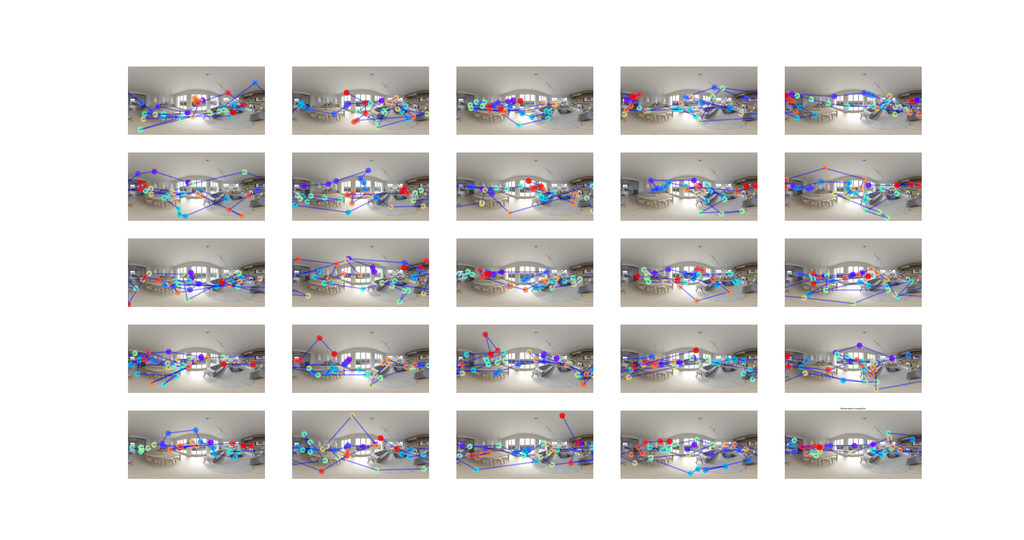}}
	\caption{Generated scanpaths for the \textit{room} scene.}
	\label{fig:ex1_2}
\end{figure*}

\begin{figure*}[t]
	\centering
	\adjustbox{trim={.1\width} {.05\height} {0.1\width} {.05\height},clip,width=\linewidth}%
  {\includegraphics[width=\linewidth]{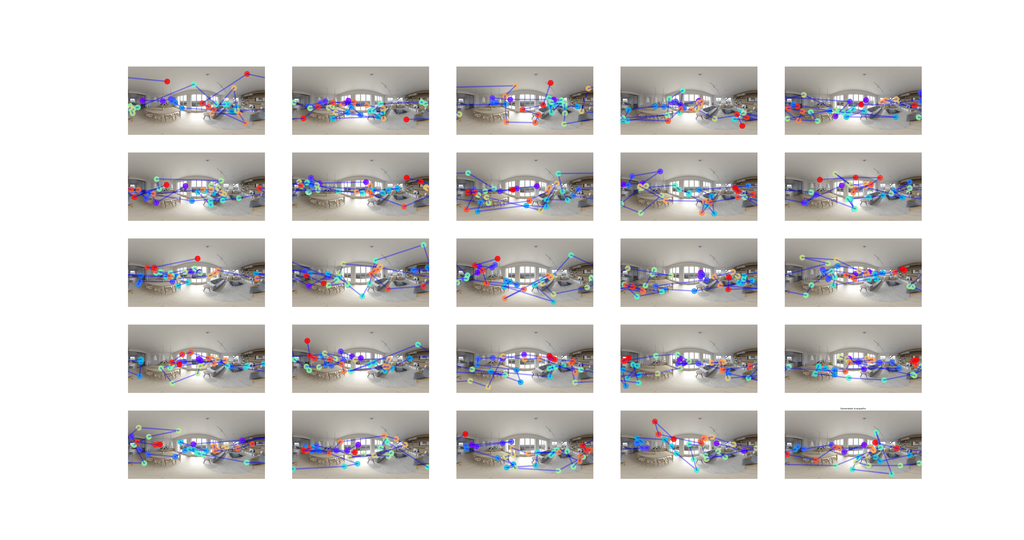}}
	\caption{Generated scanpaths for the \textit{room} scene.}
	\label{fig:ex1_3}
\end{figure*}

\begin{figure*}[t]
	\centering
	\adjustbox{trim={.1\width} {.05\height} {0.1\width} {.05\height},clip,width=\linewidth}%
  {\includegraphics[width=\linewidth]{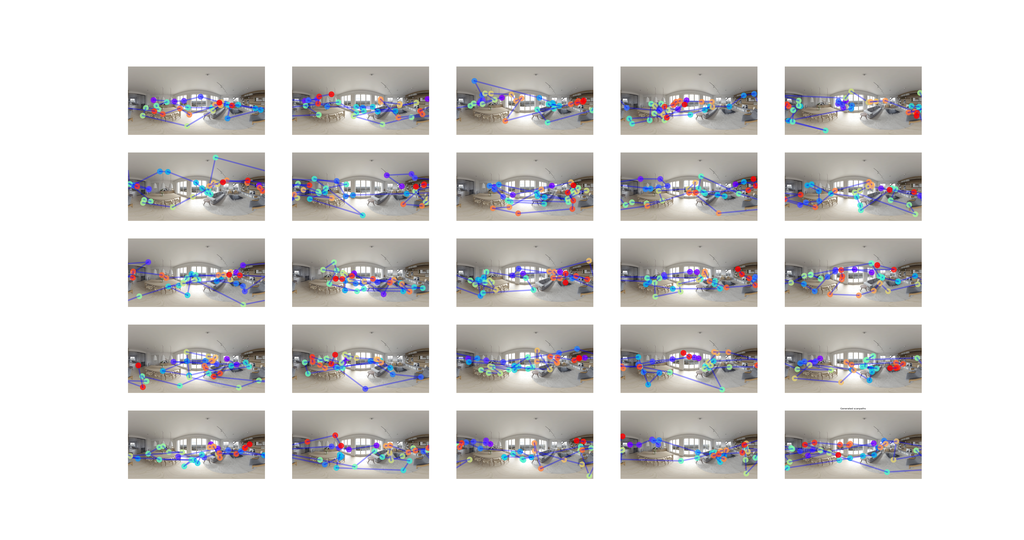}}
	\caption{Generated scanpaths for the \textit{room} scene.}
	\label{fig:ex1_4}
\end{figure*}

\begin{figure*}[t]
	\centering
	\adjustbox{trim={.1\width} {.05\height} {0.1\width} {.05\height},clip,width=\linewidth}%
  {\includegraphics[width=\linewidth]{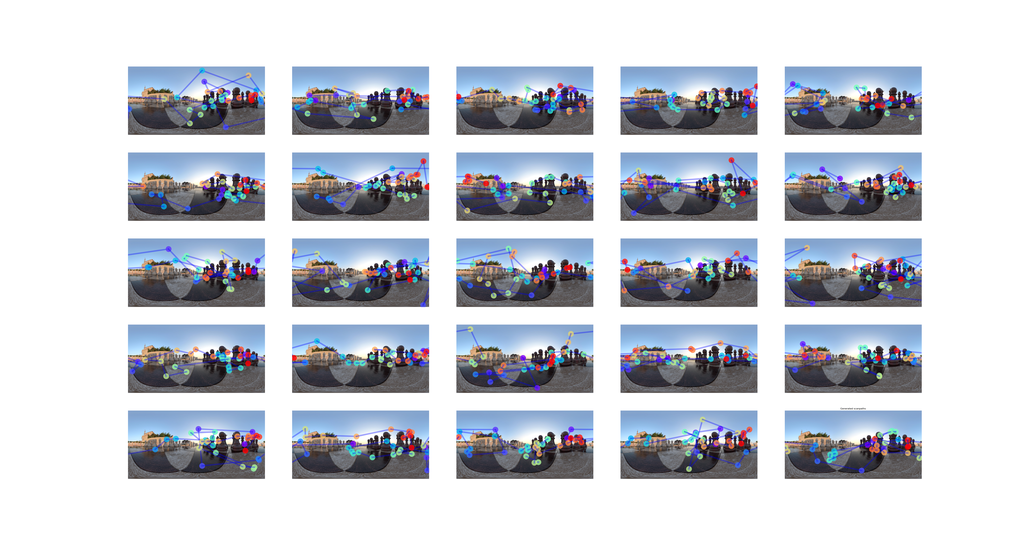}}
	\caption{Generated scanpaths for the \textit{chess} scene.}
	\label{fig:ex2_1}
\end{figure*}

\begin{figure*}[t]
	\centering
	\adjustbox{trim={.1\width} {.05\height} {0.1\width} {.05\height},clip,width=\linewidth}%
  {\includegraphics[width=\linewidth]{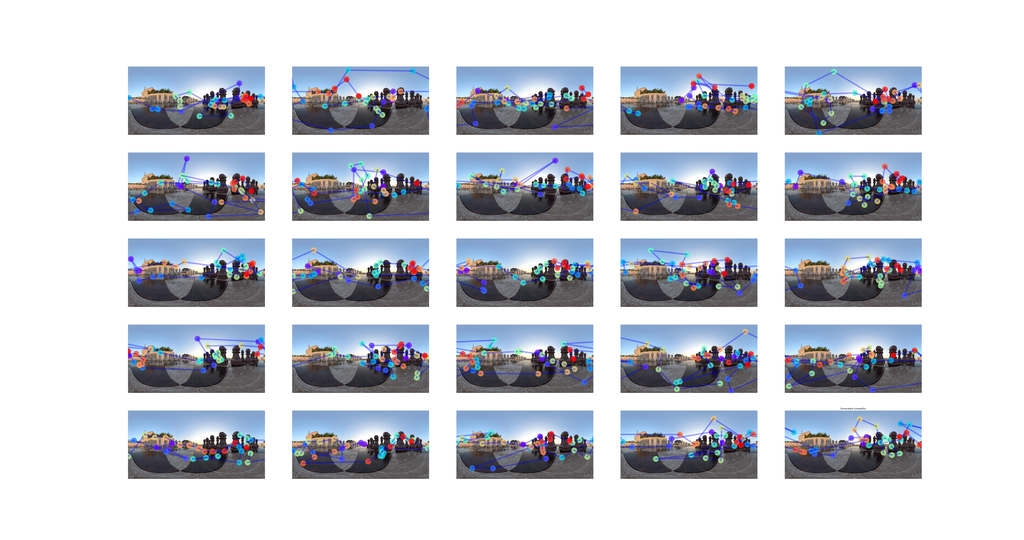}}
	\caption{Generated scanpaths for the \textit{chess} scene.}
	\label{fig:ex2_2}
\end{figure*}

\begin{figure*}[t]
	\centering
	\adjustbox{trim={.1\width} {.05\height} {0.1\width} {.05\height},clip,width=\linewidth}%
  {\includegraphics[width=\linewidth]{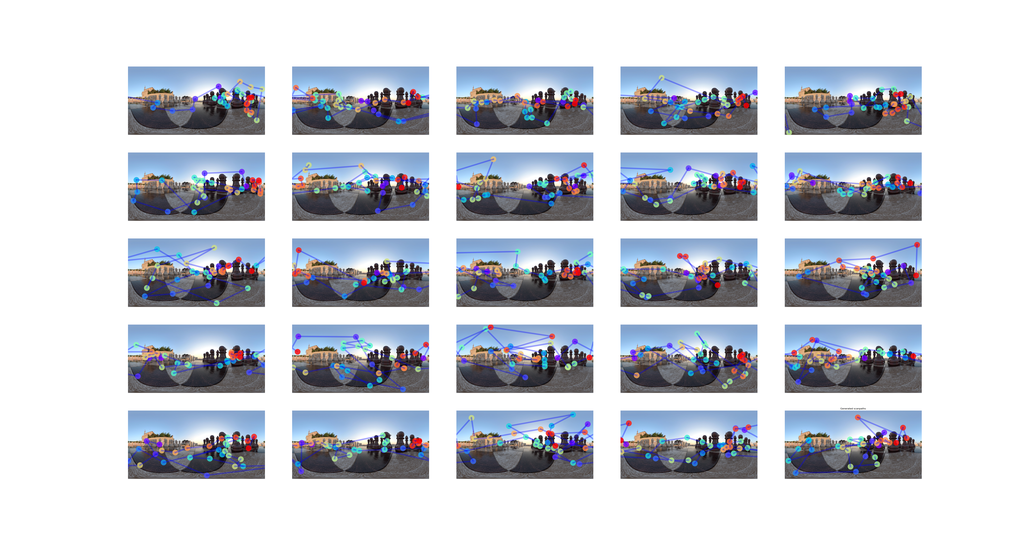}}
	\caption{Generated scanpaths for the \textit{chess} scene.}
	\label{fig:ex2_3}
\end{figure*}

\begin{figure*}[t]
	\centering
	\adjustbox{trim={.1\width} {.05\height} {0.1\width} {.05\height},clip,width=\linewidth}%
  {\includegraphics[width=\linewidth]{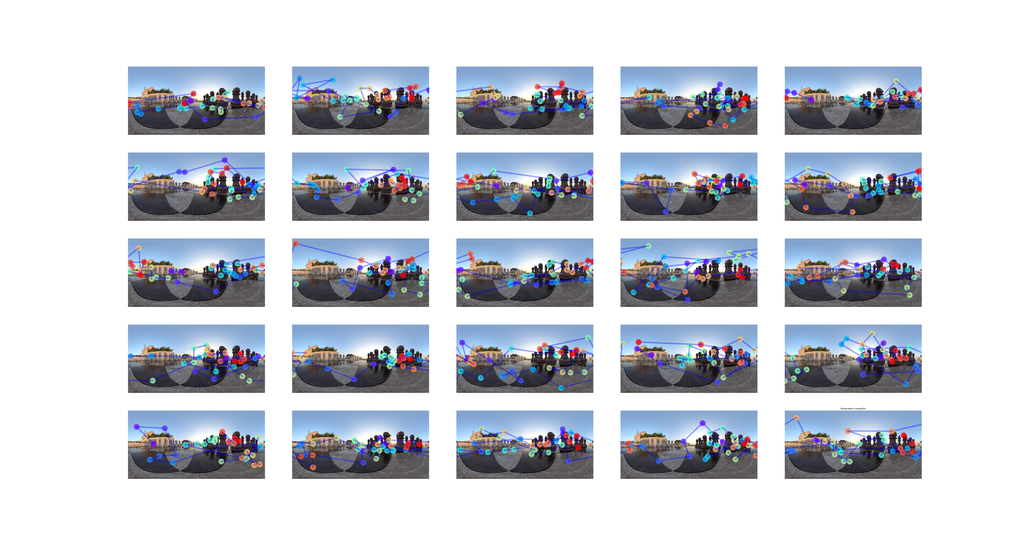}}
	\caption{Generated scanpaths for the \textit{chess} scene.}
	\label{fig:ex2_4}
\end{figure*}

\begin{figure*}[t]
	\centering
	\adjustbox{trim={.1\width} {.05\height} {0.1\width} {.05\height},clip,width=\linewidth}%
  {\includegraphics[width=\linewidth]{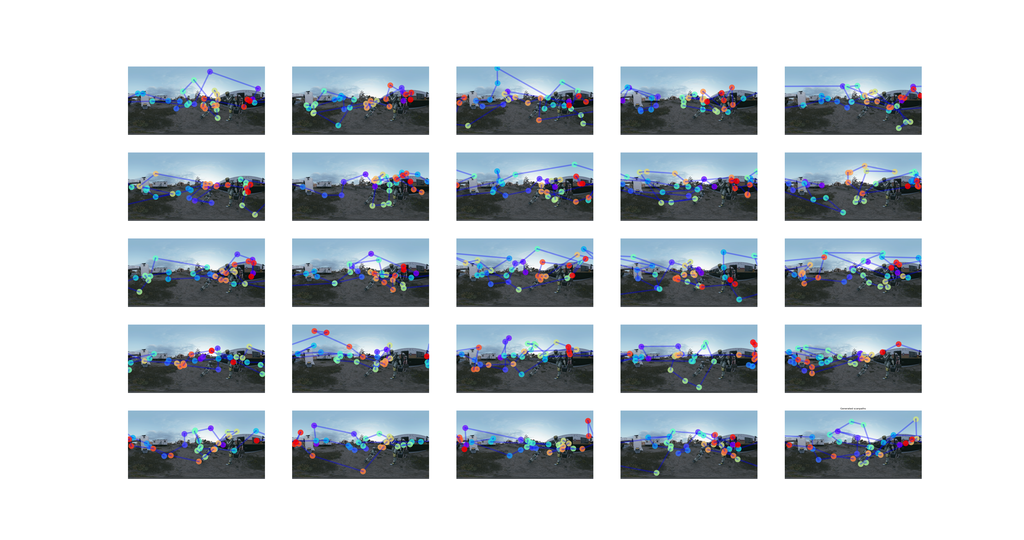}}
	\caption{Generated scanpaths for the \textit{robots} scene.}
	\label{fig:ex3_1}
\end{figure*}

\begin{figure*}[t]
	\centering
	\adjustbox{trim={.1\width} {.05\height} {0.1\width} {.05\height},clip,width=\linewidth}%
  {\includegraphics[width=\linewidth]{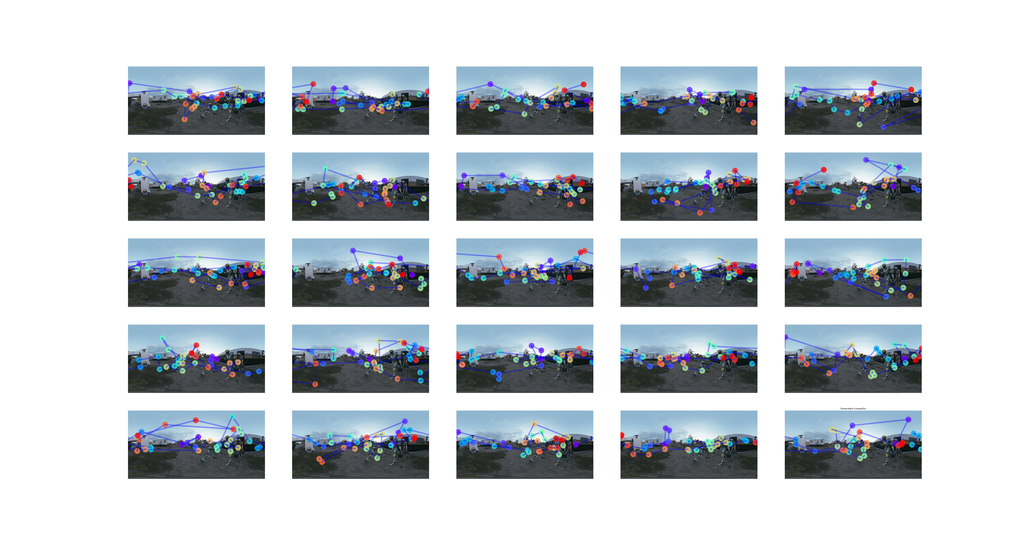}}
	\caption{Generated scanpaths for the \textit{robots} scene.}
	\label{fig:ex3_2}
\end{figure*}

\begin{figure*}[t]
	\centering
	\adjustbox{trim={.1\width} {.05\height} {0.1\width} {.05\height},clip,width=\linewidth}%
  {\includegraphics[width=\linewidth]{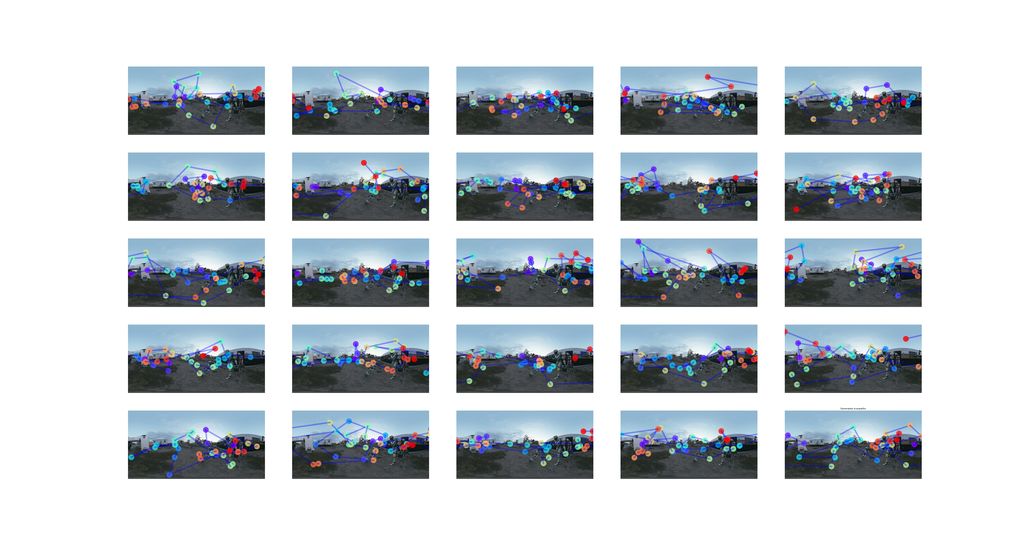}}
	\caption{Generated scanpaths for the \textit{robots} scene.}
	\label{fig:ex3_3}
\end{figure*}

\begin{figure*}[t]
	\centering
	\adjustbox{trim={.1\width} {.05\height} {0.1\width} {.05\height},clip,width=\linewidth}%
  {\includegraphics[width=\linewidth]{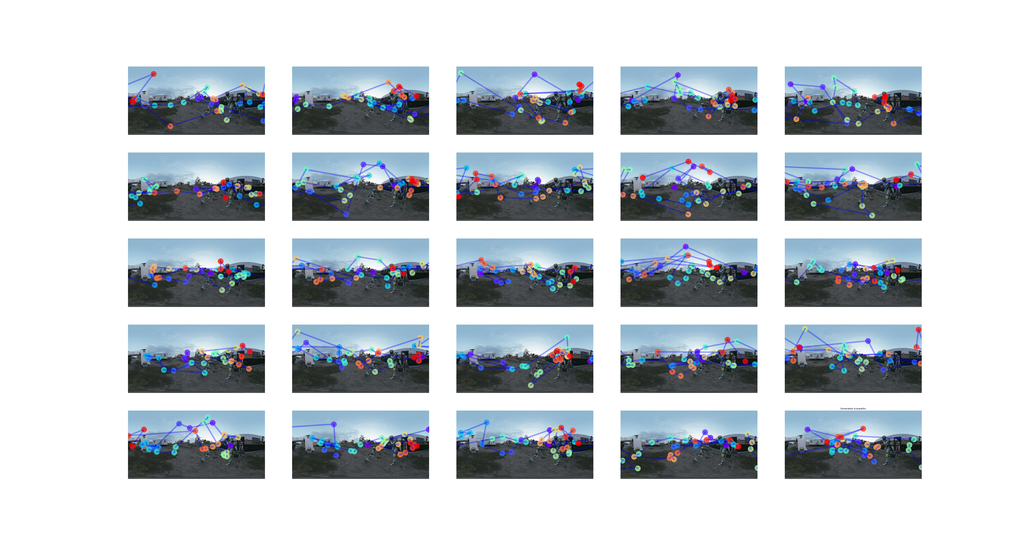}}
	\caption{Generated scanpaths for the \textit{robots} scene.}
	\label{fig:ex3_4}
\end{figure*}

\begin{figure*}[t]
	\centering
   \adjustbox{trim={.1\width} {.05\height} {0.1\width} {.05\height},clip,width=\linewidth}%
  {\includegraphics[width=\linewidth]{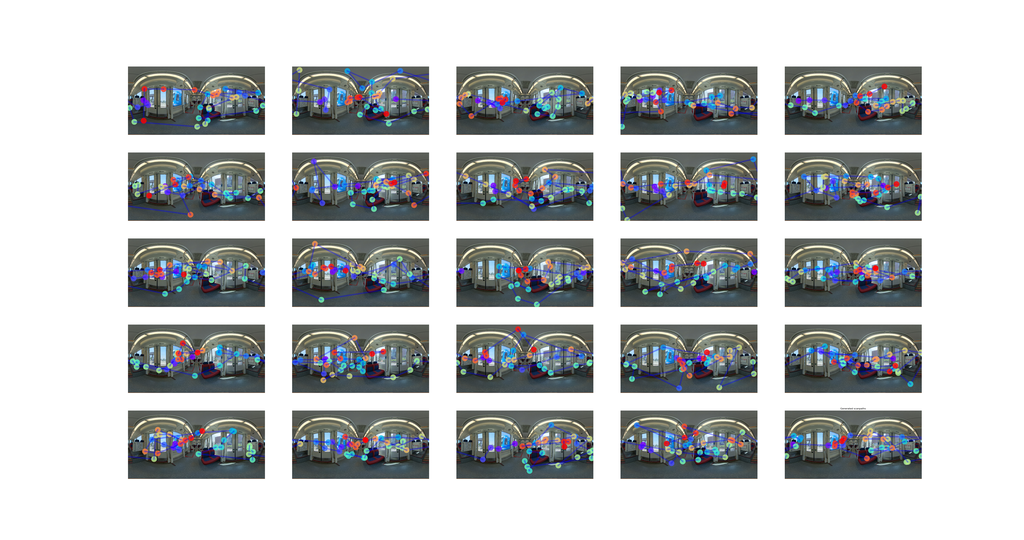}}
	\caption{Generated scanpaths for the \textit{train} scene.}
	\label{fig:ex4_1}
\end{figure*}

\begin{figure*}[t]
	\centering
	\adjustbox{trim={.1\width} {.05\height} {0.1\width} {.05\height},clip,width=\linewidth}%
  {\includegraphics[width=\linewidth]{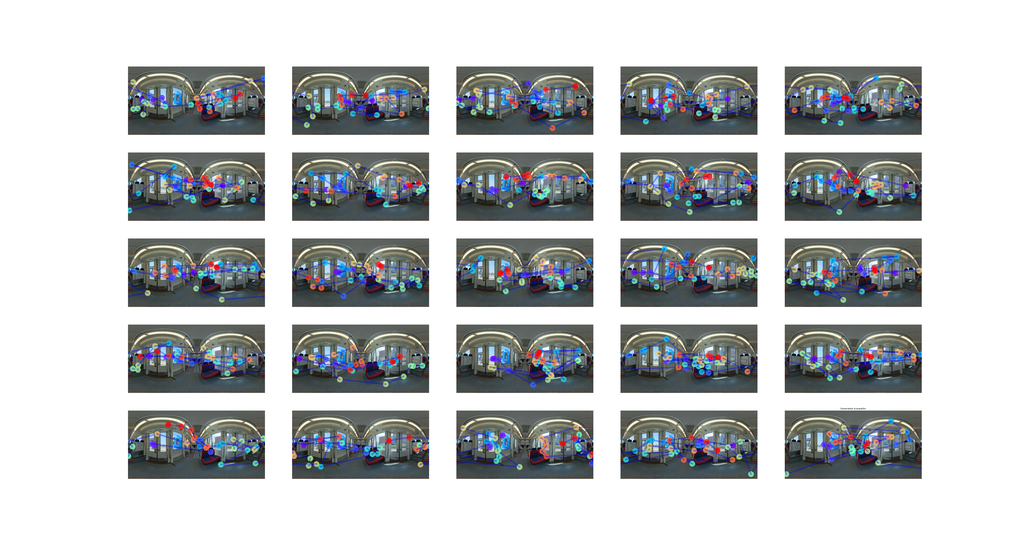}}
	\caption{Generated scanpaths for the \textit{train} scene.}
	\label{fig:ex4_2}
\end{figure*}

\begin{figure*}[t]
	\centering
	\adjustbox{trim={.1\width} {.05\height} {0.1\width} {.05\height},clip,width=\linewidth}%
  {\includegraphics[width=\linewidth]{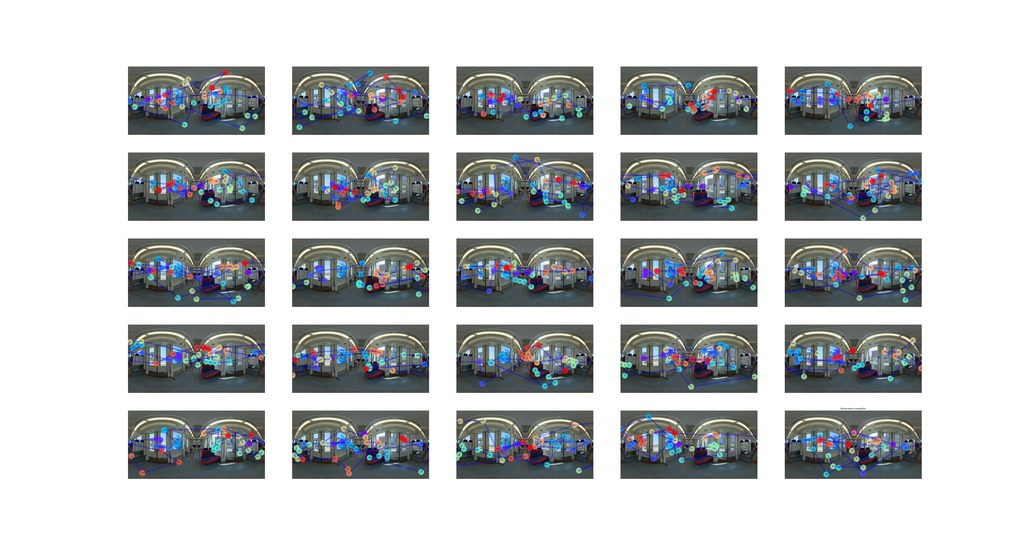}}
	\caption{Generated scanpaths for the \textit{train} scene.}
	\label{fig:ex4_3}
\end{figure*}

\begin{figure*}[t]
	\centering
	\adjustbox{trim={.1\width} {.05\height} {0.1\width} {.05\height},clip,width=\linewidth}%
  {\includegraphics[width=\linewidth]{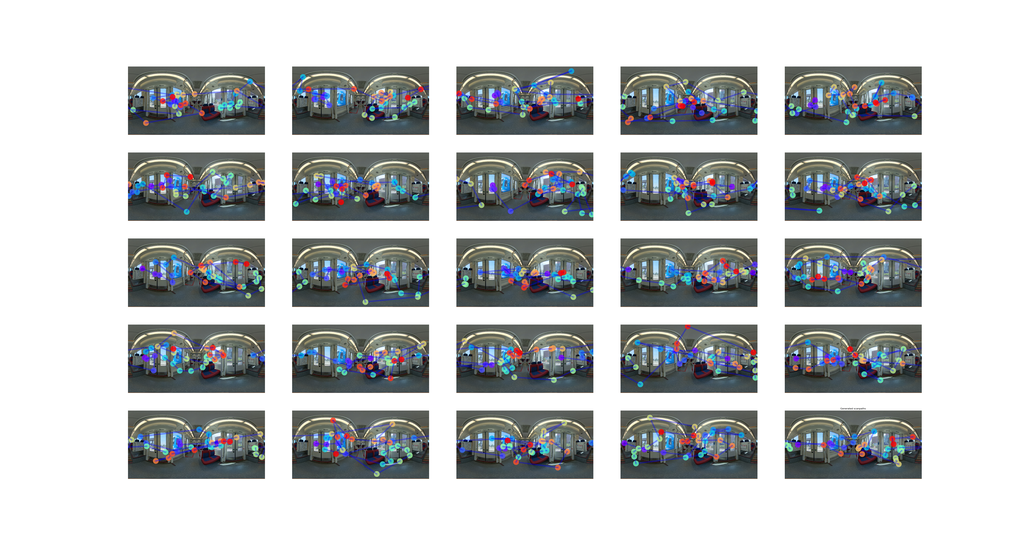}}
	\caption{Generated scanpaths for the \textit{train} scene.}
	\label{fig:ex4_4}
\end{figure*}

\begin{figure*}[t]
	\centering
	\adjustbox{trim={.1\width} {.05\height} {0.1\width} {.05\height},clip,width=\linewidth}%
  {\includegraphics[width=\linewidth]{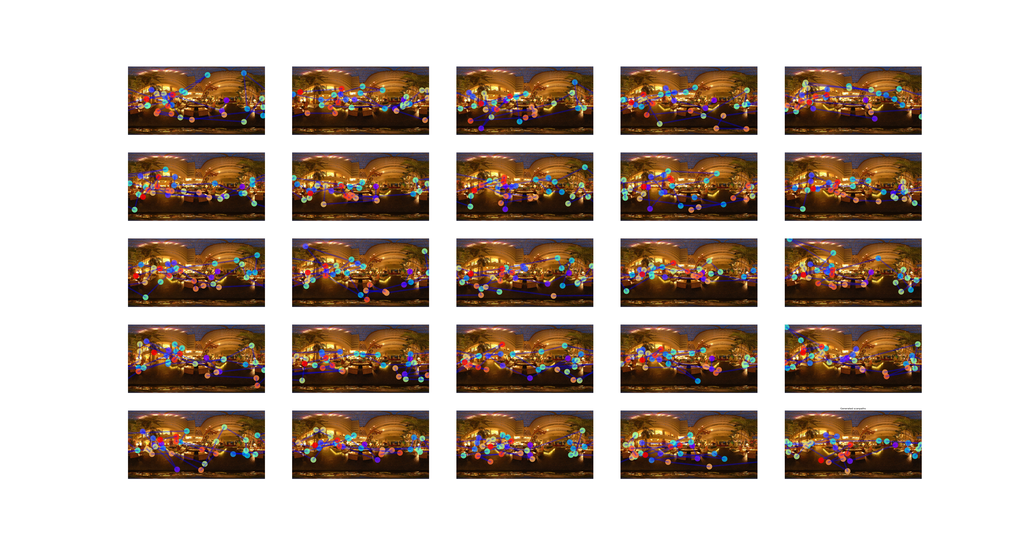}}
	\caption{Generated scanpaths for the \textit{resort} scene.}
	\label{fig:ex5_1}
\end{figure*}

\begin{figure*}[t]
	\centering
	\adjustbox{trim={.1\width} {.05\height} {0.1\width} {.05\height},clip,width=\linewidth}%
  {\includegraphics[width=\linewidth]{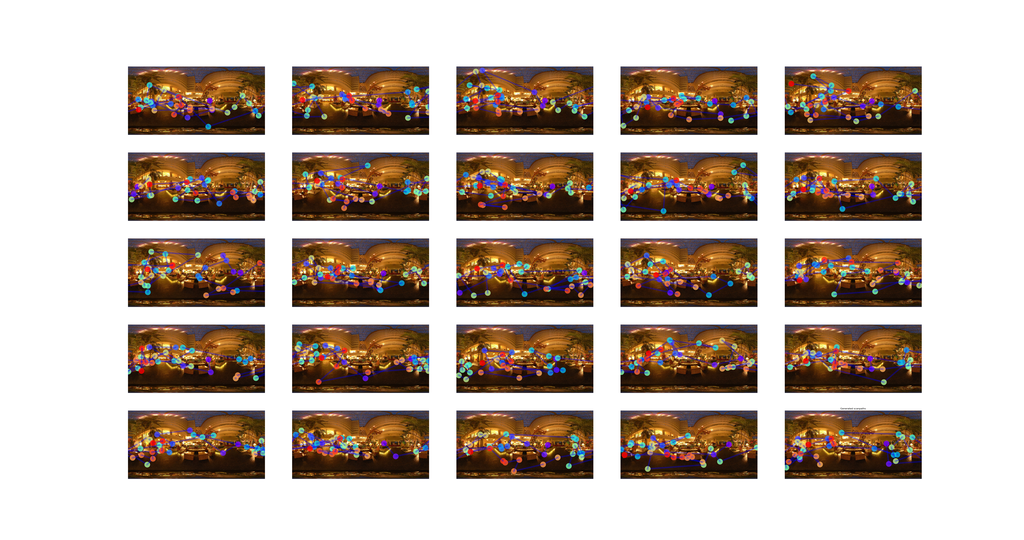}}
	\caption{Generated scanpaths for the \textit{resort} scene.}
	\label{fig:ex5_2}
\end{figure*}

\begin{figure*}[t]
	\centering
	\adjustbox{trim={.1\width} {.05\height} {0.1\width} {.05\height},clip,width=\linewidth}%
  {\includegraphics[width=\linewidth]{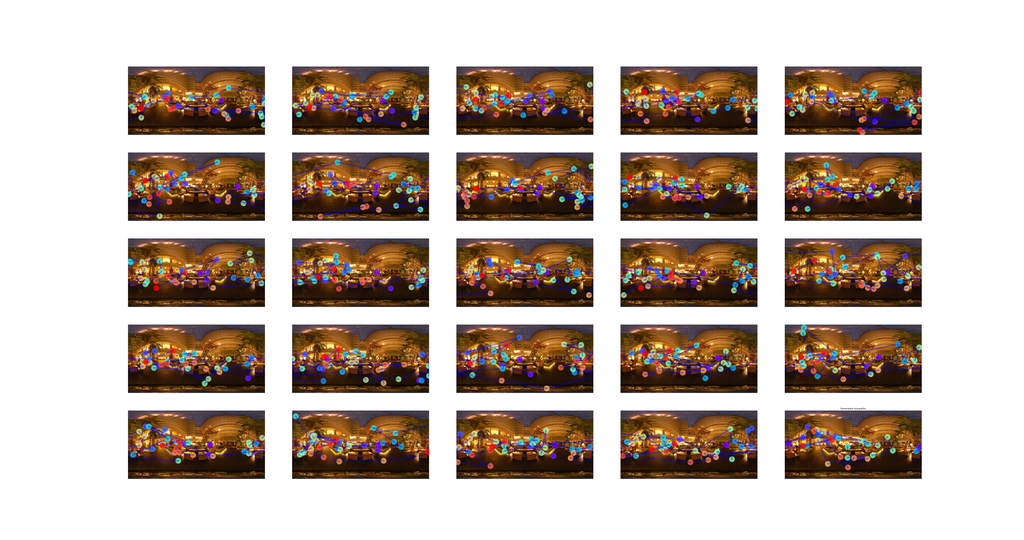}}
	\caption{Generated scanpaths for the \textit{resort} scene.}
	\label{fig:ex5_3}
\end{figure*}

\begin{figure*}[t]
	\centering
	\adjustbox{trim={.1\width} {.05\height} {0.1\width} {.05\height},clip,width=\linewidth}%
  {\includegraphics[width=\linewidth]{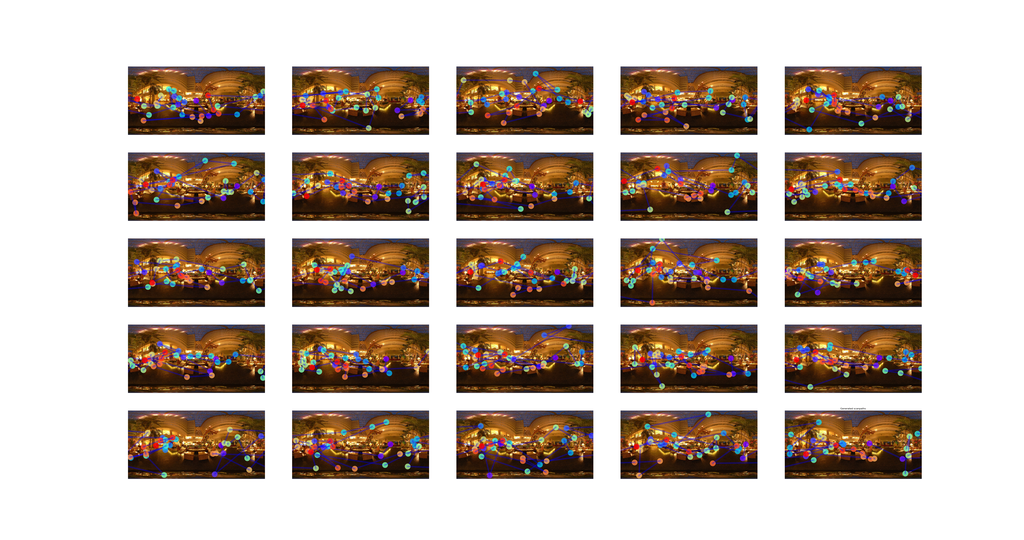}}
	\caption{Generated scanpaths for the \textit{resort} scene.}
	\label{fig:ex5_4}
\end{figure*}

\begin{figure*}[t]
	\centering
	\adjustbox{trim={.1\width} {.05\height} {0.1\width} {.05\height},clip,width=\linewidth}%
  {\includegraphics[width=\linewidth]{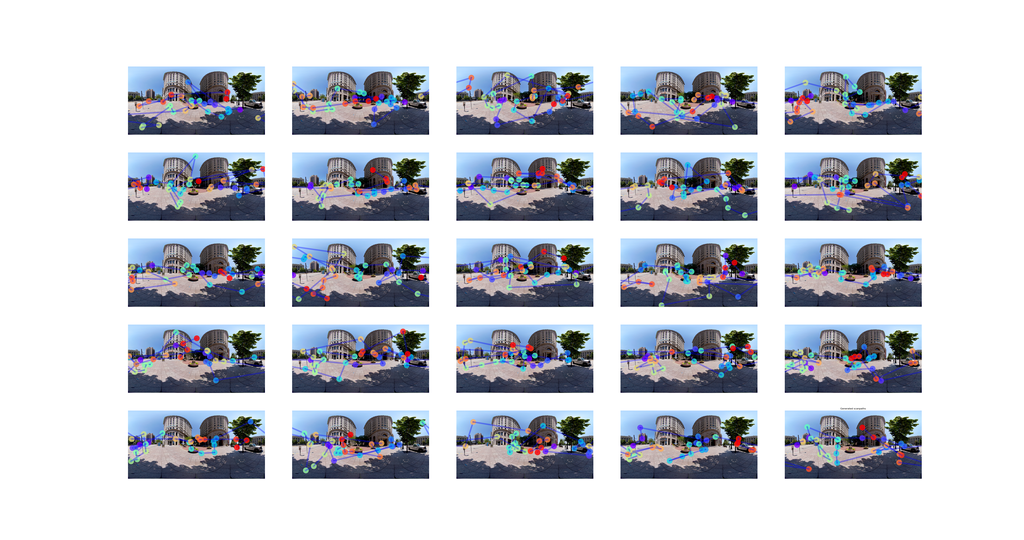}}
	\caption{Generated scanpaths for the \textit{square} scene.}
	\label{fig:ex6_1}
\end{figure*}

\begin{figure*}[t]
	\centering
	\adjustbox{trim={.1\width} {.05\height} {0.1\width} {.05\height},clip,width=\linewidth}%
  {\includegraphics[width=\linewidth]{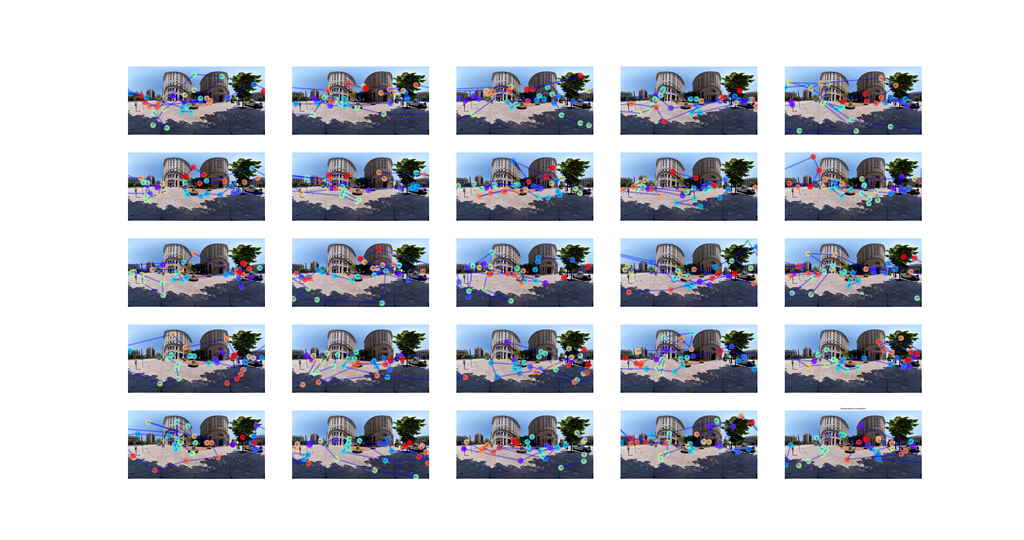}}
	\caption{Generated scanpaths for the \textit{square} scene.}
	\label{fig:ex6_2}
\end{figure*}

\begin{figure*}[t]
	\centering
	\adjustbox{trim={.1\width} {.05\height} {0.1\width} {.05\height},clip,width=\linewidth}%
  {\includegraphics[width=\linewidth]{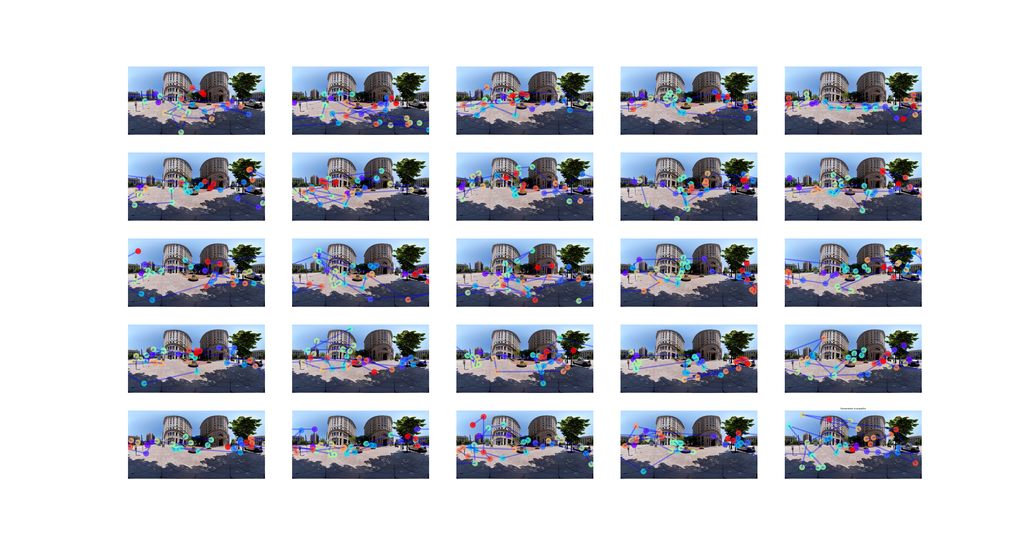}}
	\caption{Generated scanpaths for the \textit{square} scene.}
	\label{fig:ex6_3}
\end{figure*}

\begin{figure*}[t]
	\centering
	\adjustbox{trim={.1\width} {.05\height} {0.1\width} {.05\height},clip,width=\linewidth}%
  {\includegraphics[width=\linewidth]{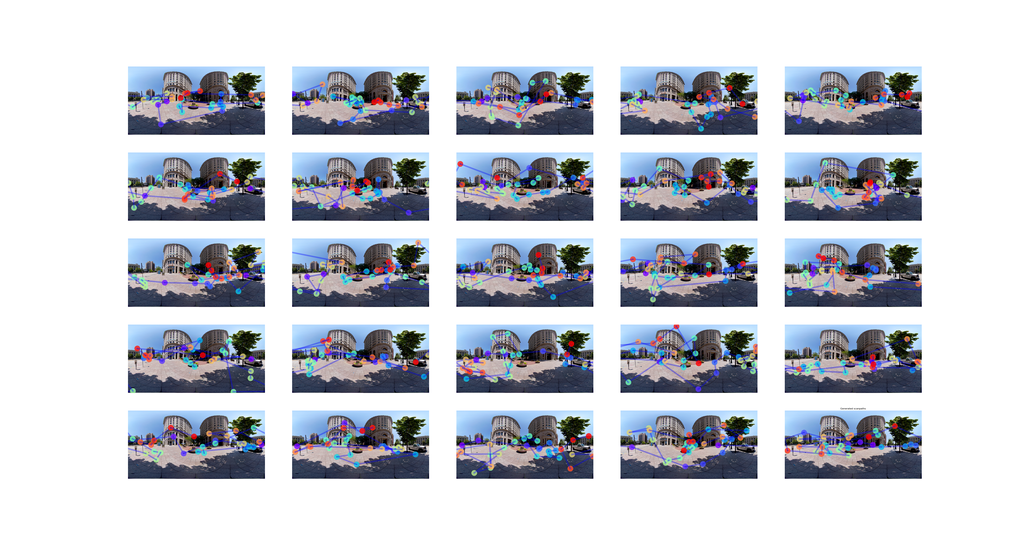}}
	\caption{Generated scanpaths for the \textit{square} scene.}
	\label{fig:ex6_4}
\end{figure*}

\begin{figure*}[t]
	\centering
	\adjustbox{trim={.1\width} {.05\height} {0.1\width} {.05\height},clip,width=\linewidth}%
  {\includegraphics[width=\linewidth]{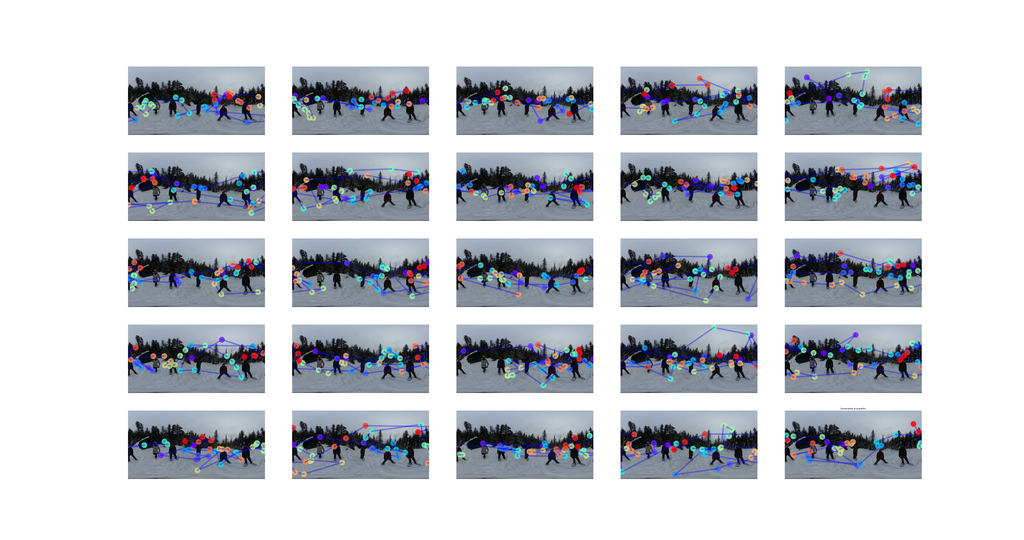}}
	\caption{Generated scanpaths for the \textit{snow} scene.}
	\label{fig:ex7_1}
\end{figure*}

\begin{figure*}[t]
	\centering
	\adjustbox{trim={.1\width} {.05\height} {0.1\width} {.05\height},clip,width=\linewidth}%
  {\includegraphics[width=\linewidth]{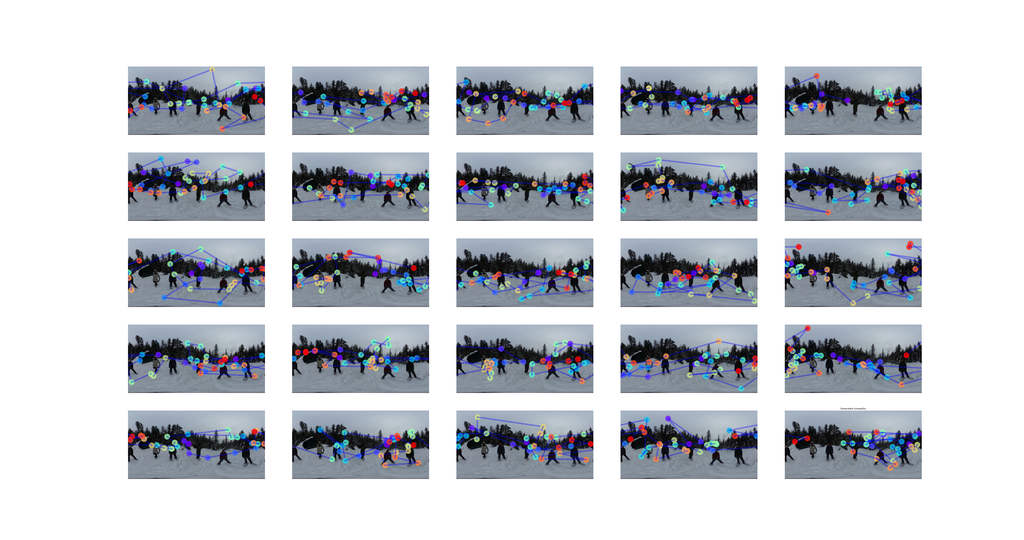}}
	\caption{Generated scanpaths for the \textit{snow} scene.}
	\label{fig:ex7_2}
\end{figure*}

\begin{figure*}[t]
	\centering
	\adjustbox{trim={.1\width} {.05\height} {0.1\width} {.05\height},clip,width=\linewidth}%
  {\includegraphics[width=\linewidth]{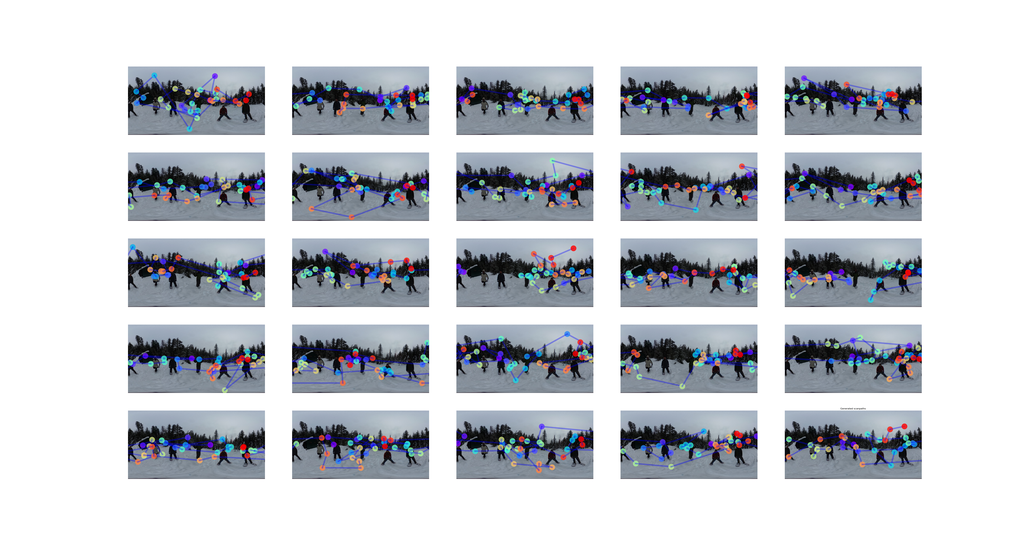}}
	\caption{Generated scanpaths for the \textit{snow} scene.}
	\label{fig:ex7_3}
\end{figure*}

\begin{figure*}[t]
	\centering
	\adjustbox{trim={.1\width} {.05\height} {0.1\width} {.05\height},clip,width=\linewidth}%
  {\includegraphics[width=\linewidth]{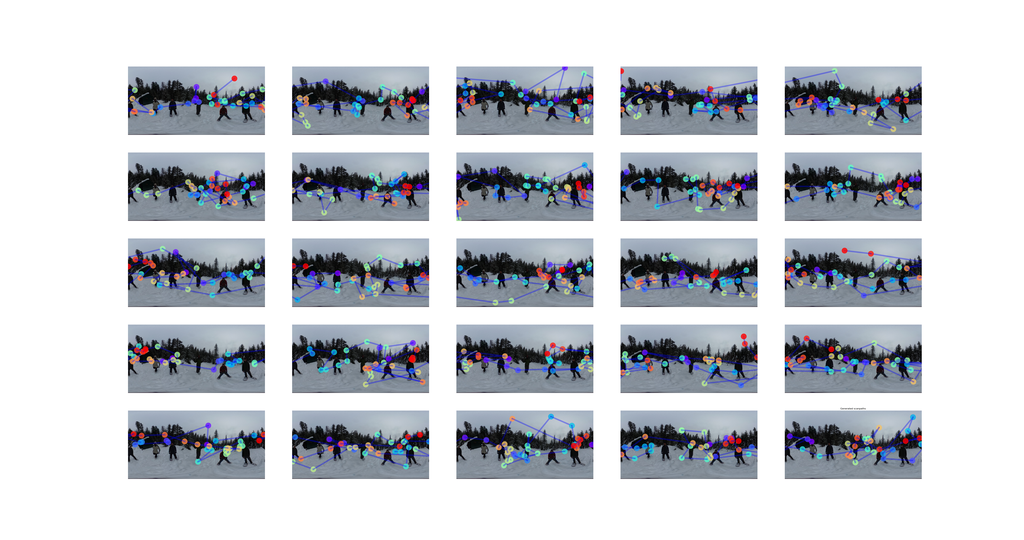}}
	\caption{Generated scanpaths for the \textit{snow} scene.}
	\label{fig:ex7_4}
\end{figure*}

\begin{figure*}[t]
	\centering
	\adjustbox{trim={.1\width} {.05\height} {0.1\width} {.05\height},clip,width=\linewidth}%
  {\includegraphics[width=\linewidth]{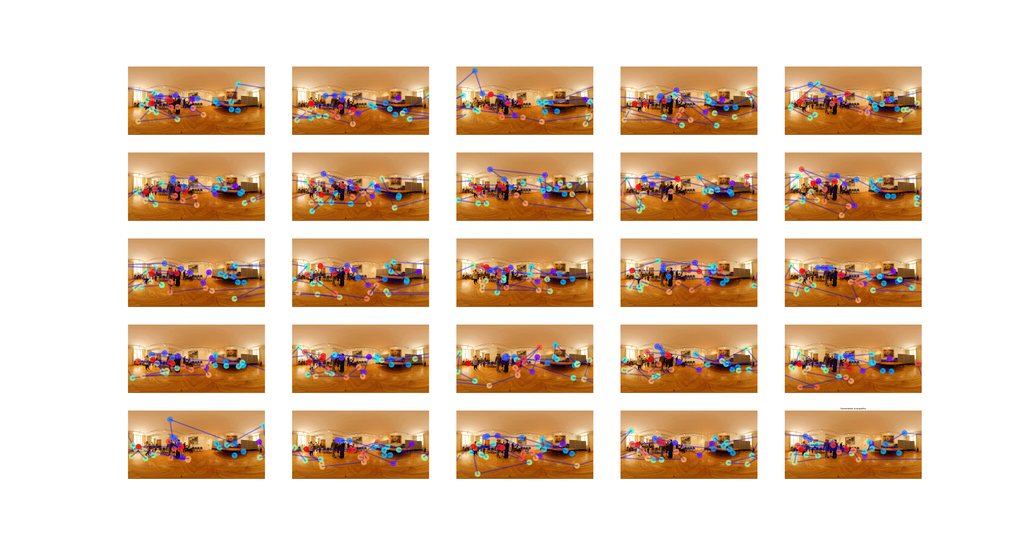}}
	\caption{Generated scanpaths for the \textit{museum} scene.}
	\label{fig:ex8_1}
\end{figure*}

\begin{figure*}[t]
	\centering
	\adjustbox{trim={.1\width} {.05\height} {0.1\width} {.05\height},clip,width=\linewidth}%
  {\includegraphics[width=\linewidth]{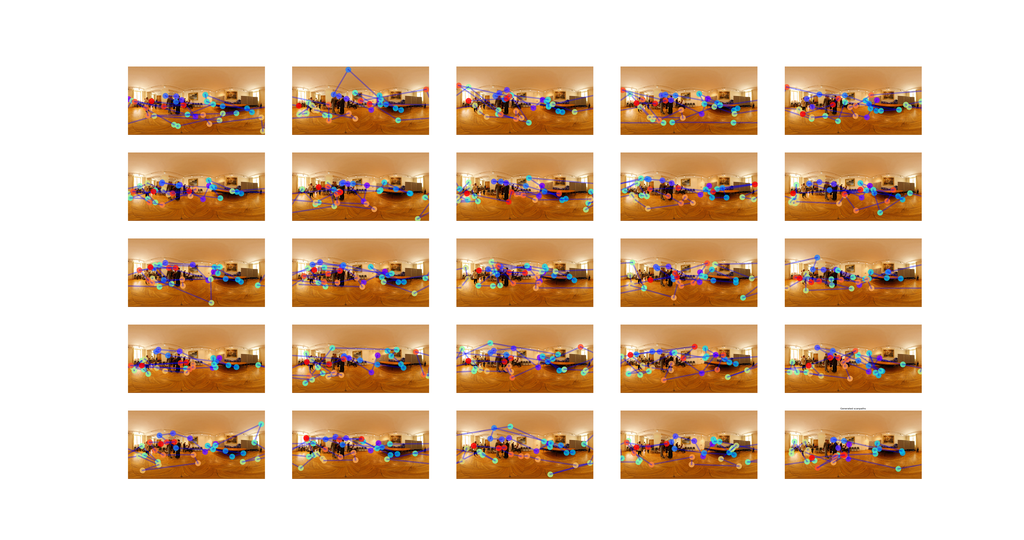}}
	\caption{Generated scanpaths for the \textit{museum} scene.}
	\label{fig:ex8_2}
\end{figure*}

\begin{figure*}[t]
	\centering
	\adjustbox{trim={.1\width} {.05\height} {0.1\width} {.05\height},clip,width=\linewidth}%
  {\includegraphics[width=\linewidth]{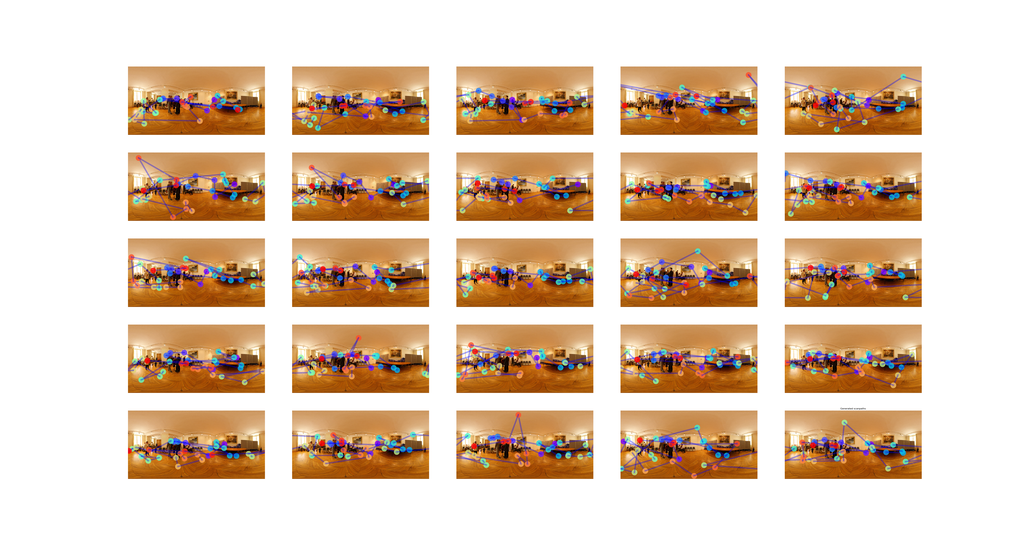}}
	\caption{Generated scanpaths for the \textit{museum} scene.}
	\label{fig:ex8_3}
\end{figure*}

\begin{figure*}[t]
	\centering
	\adjustbox{trim={.1\width} {.05\height} {0.1\width} {.05\height},clip,width=\linewidth}%
  {\includegraphics[width=\linewidth]{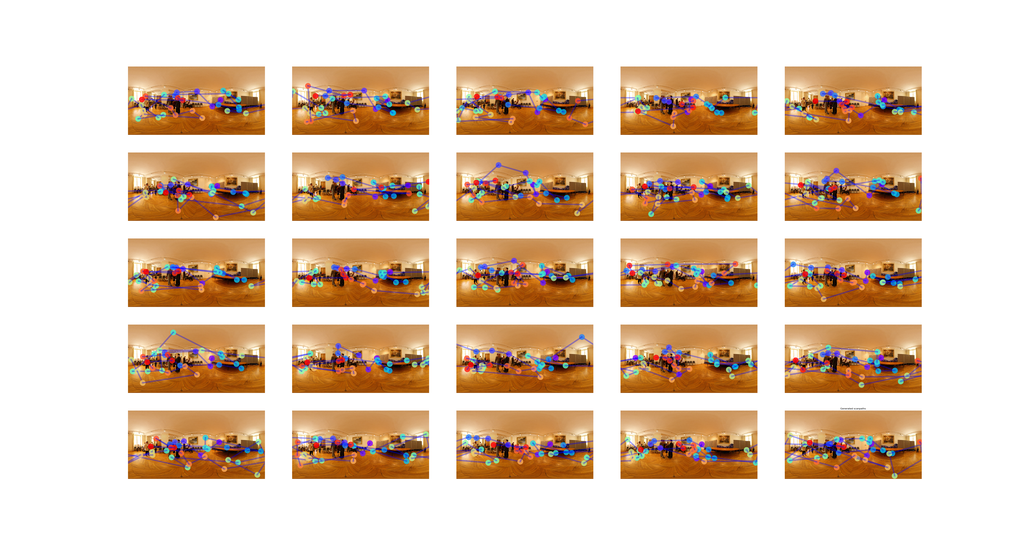}}
	\caption{Generated scanpaths for the \textit{museum} scene.}
	\label{fig:ex8_4}
\end{figure*}

\begin{figure*}[t]
	\centering
		\adjustbox{trim={.1\width} {.05\height} {0.1\width} {.05\height},clip,width=\linewidth}%
  {\includegraphics[width=\linewidth]{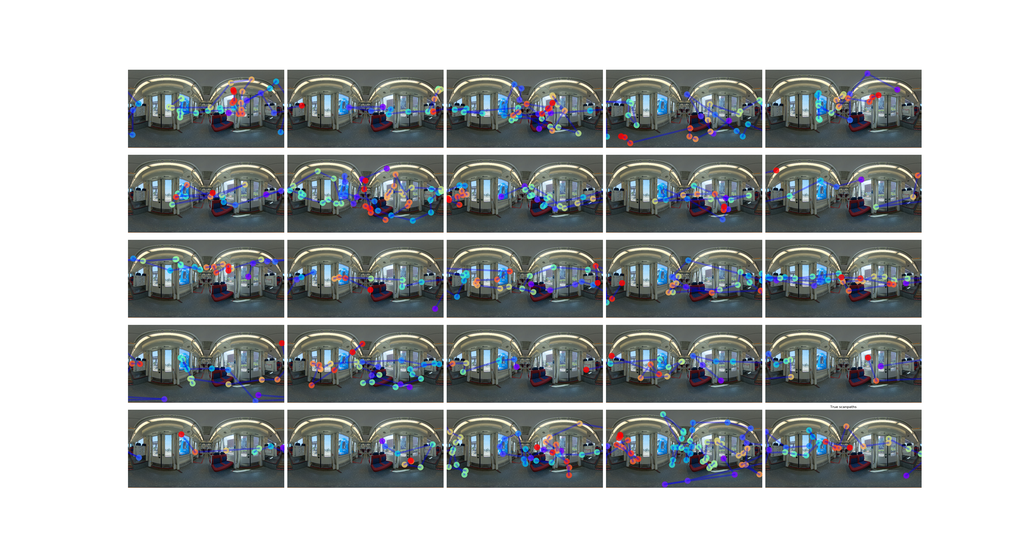}}
	\caption{Ground truth scanpaths for the \textit{train} scene.}
	\label{fig:supp_gt_comp_1}
\end{figure*}

\begin{figure*}[t]
	\centering
		\adjustbox{trim={.1\width} {.05\height} {0.1\width} {.05\height},clip,width=\linewidth}%
  {\includegraphics[width=\linewidth]{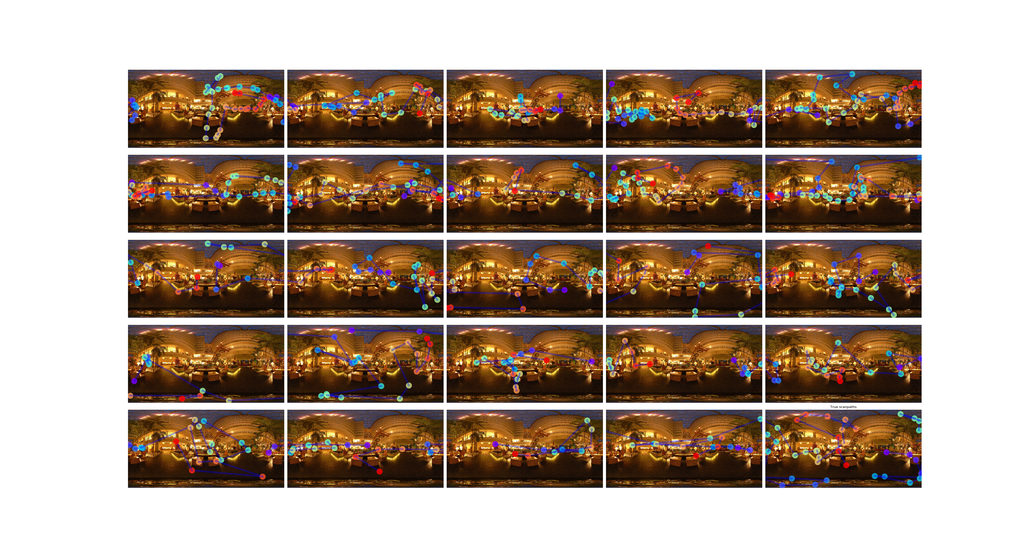}}
	\caption{Ground truth scanpaths for the \textit{resort} scene.}
	\label{fig:supp_gt_comp_2}
\end{figure*}

\begin{figure*}[t]
	\centering
		\adjustbox{trim={.1\width} {.05\height} {0.1\width} {.05\height},clip,width=\linewidth}%
  {\includegraphics[width=\linewidth]{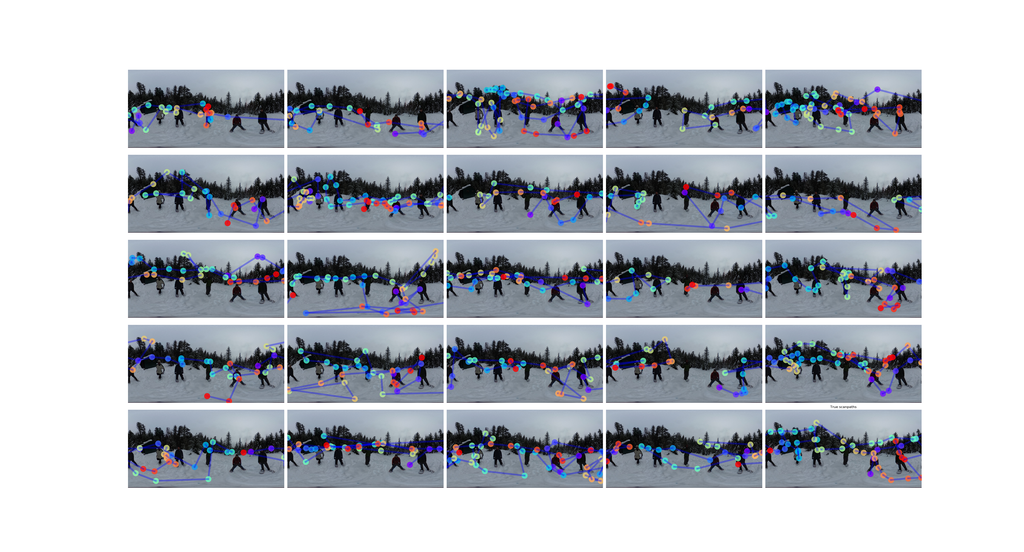}}
	\caption{Ground truth scanpaths for the \textit{snow} scene.}
	\label{fig:supp_gt_comp_3}
\end{figure*}

\begin{figure*}[t]
	\centering
		\adjustbox{trim={.1\width} {.05\height} {0.1\width} {.05\height},clip,width=\linewidth}%
  {\includegraphics[width=\linewidth]{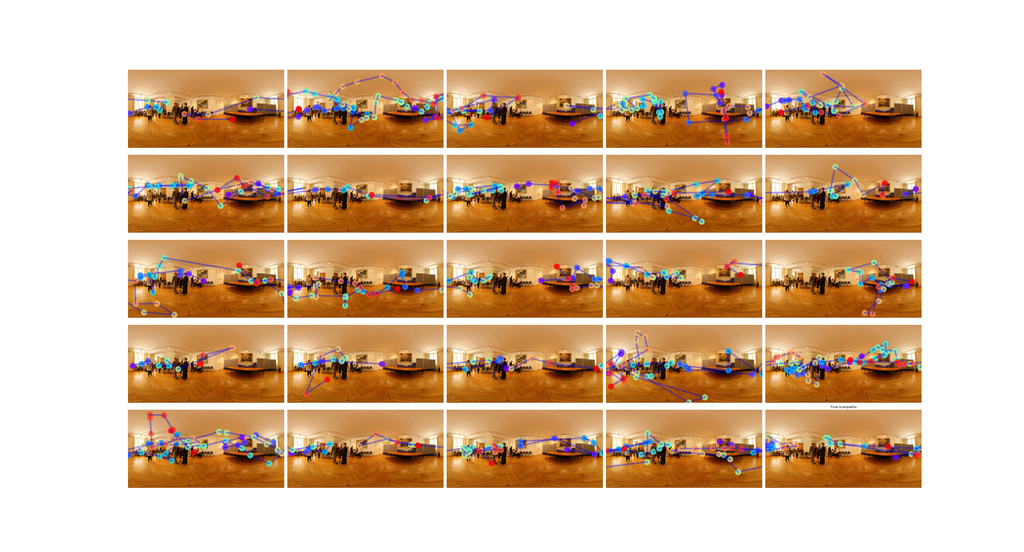}}
	\caption{Ground truth scanpaths for the \textit{museum} scene.}
	\label{fig:supp_gt_comp_4}
\end{figure*}

\begin{figure*}[t]
	\centering
		\adjustbox{trim={.1\width} {.05\height} {0.1\width} {.05\height},clip,width=\linewidth}%
  {\includegraphics[width=\linewidth]{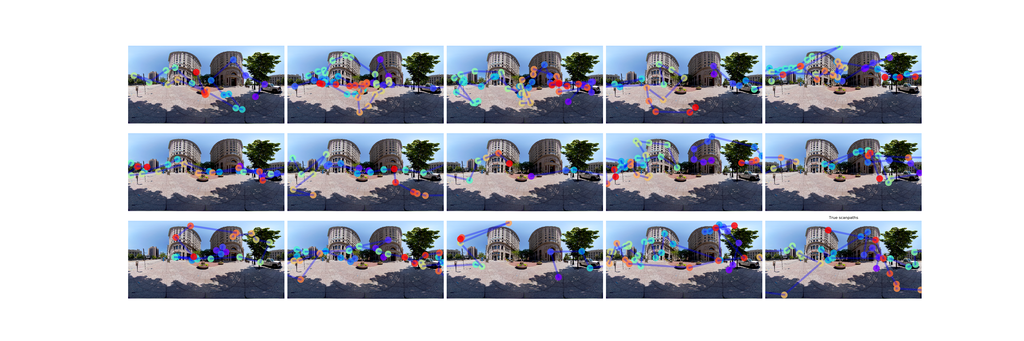}}
	\caption{Ground truth scanpaths for the \textit{square} scene.}
	\label{fig:supp_gt_comp_5}
\end{figure*}

\begin{figure*}[t]
	\centering
		\adjustbox{trim={.1\width} {.05\height} {0.1\width} {.05\height},clip,width=\linewidth}%
  {\includegraphics[width=\linewidth]{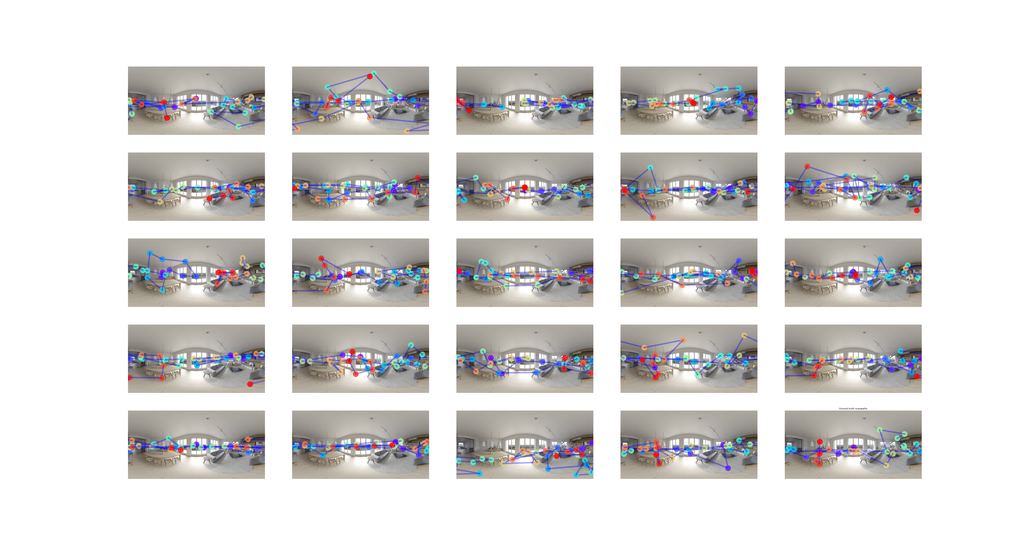}}
	\caption{Ground truth scanpaths for the \textit{room} scene.}
	\label{fig:gt_sitz_1}
\end{figure*}

\begin{figure*}[t]
	\centering
		\adjustbox{trim={.1\width} {.05\height} {0.1\width} {.05\height},clip,width=\linewidth}%
  {\includegraphics[width=\linewidth]{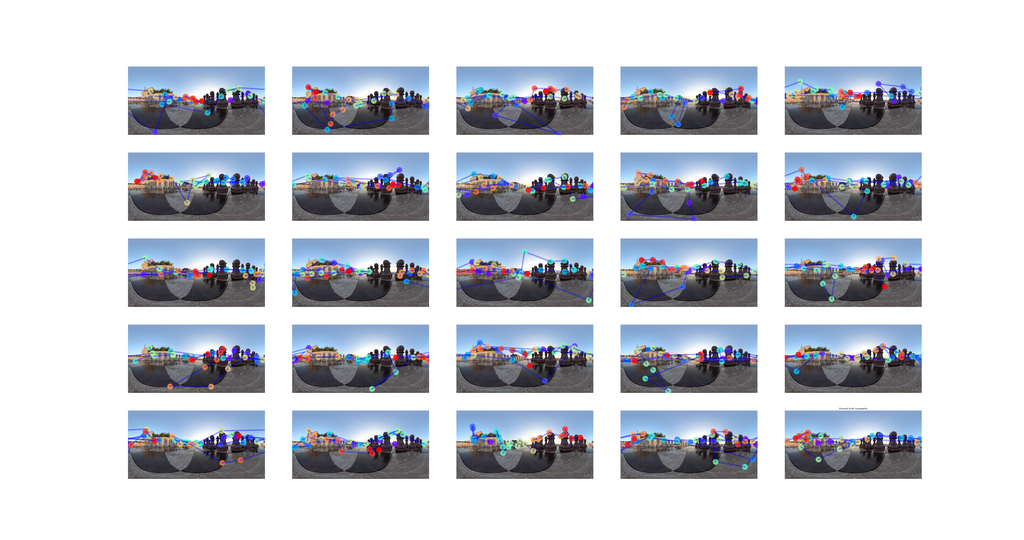}}
	\caption{Ground truth scanpaths for the \textit{chess} scene.}
	\label{fig:gt_sitz_2}
\end{figure*}

\begin{figure*}[t]
	\centering
		\adjustbox{trim={.1\width} {.05\height} {0.1\width} {.05\height},clip,width=\linewidth}%
  {\includegraphics[width=\linewidth]{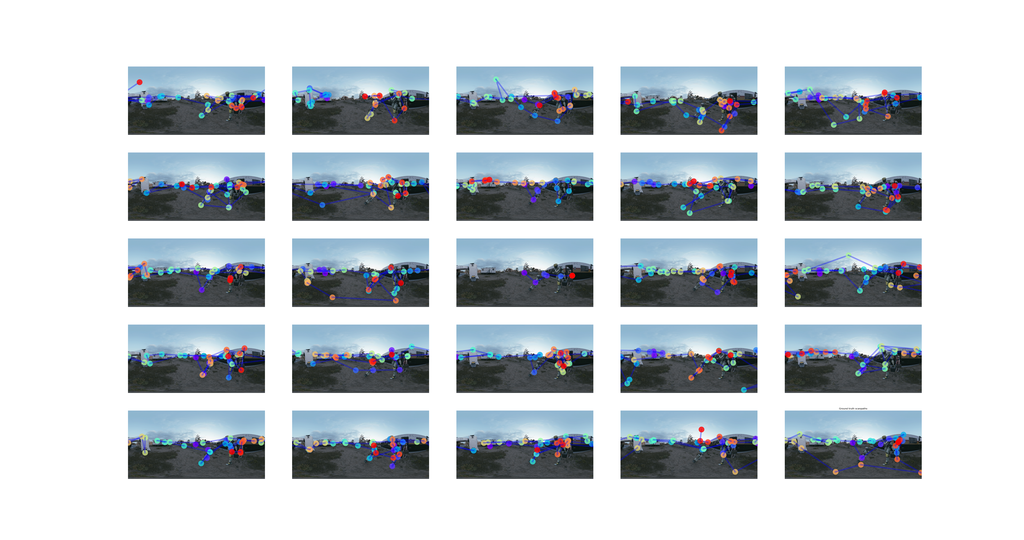}}
	\caption{Ground truth scanpaths for the \textit{robots} scene.}
	\label{fig:gt_sitz_3}
\end{figure*}

Figure~\ref{fig:app_retargeting} shows several representative steps of this process for two different scenes (\emph{chess} and \emph{street}). Full videos of several scenes are included in the supplementary video. The generated Ken Burns--style panorama previews look like a human observer exploring these panorama and provide a very intuitive preview of the complex scenes they depict.

\paragraph{Other applications}

Our model has the potential to enable other applications beyond what we have shown in this section. One such example is \emph{gaze simulation for virtual avatars}. 
When displaying or interacting with virtual characters, eye gaze is one of the most critical, yet most difficult, aspects to simulate~\cite{ruhland2015review}. Accurately simulating gaze behavior not only aids in conveying realism, but can also provide additional information such as signalling interest, aiding the conversation through non-verbal cues, facilitating turn-taking in multi-party conversations, or indicating attentiveness, among others. 
Given an avatar immersed within a virtual scene, generating plausible scanpaths conditioned by a 360$^\circ$ image of their environment could be an efficient, affordable way of driving the avatar's gaze behavior in a realistic manner.

Another potential application of our model is its use for \emph{gaze-contingent rendering}. These approaches have been proposed to save rendering time and bandwidth in VR systems or drive the user's accommodation. Eye trackers are required for these applications, but they are often too slow, making computationally efficient approaches for predicting gaze trajectories or landing positions important~\cite{arabadzhiyska2017saccade}. Our method for generating scanpaths could not only help prototype and evaluate such systems in simulation, without the need for a physical eye tracker and actual users, but also in optimizing their latency and performance during runtime.

\FloatBarrier

{\small
\bibliographystyle{ieee_fullname}
\bibliography{references}

\begin{thebibliography}{10}\itemsep=-1pt

\bibitem{arabadzhiyska2017saccade}
Elena Arabadzhiyska, Okan~Tarhan Tursun, Karol Myszkowski, Hans-Peter Seidel,
  and Piotr Didyk.
\newblock Saccade landing position prediction for gaze-contingent rendering.
\newblock {\em ACM Transactions on Graphics (TOG)}, 36(4):1--12, 2017.

\bibitem{assens2017saltinet}
Marc Assens, Xavier Giro-i Nieto, Kevin McGuinness, and Noel~E O'Connor.
\newblock Saltinet: Scan-path prediction on 360 degree images using saliency
  volumes.
\newblock In {\em Proceedings of the IEEE ICCV Workshops}, pages 2331--2338,
  2017.

\bibitem{assens2018pathgan}
Marc Assens, Xavier Giro-i Nieto, Kevin McGuinness, and Noel~E O'Connor.
\newblock Pathgan: visual scanpath prediction with generative adversarial
  networks.
\newblock In {\em Proceedings of the European Conference on Computer Vision
  (ECCV)}, pages 0--0, 2018.

\bibitem{assens2018scanpath}
Marc Assens, Xavier Giro-i Nieto, Kevin McGuinness, and Noel~E O’Connor.
\newblock Scanpath and saliency prediction on 360 degree images.
\newblock {\em Signal Processing: Image Communication}, 69:8--14, 2018.

\bibitem{bao2020scanpath}
Wentao Bao and Zhenzhong Chen.
\newblock Human scanpath prediction based on deep convolutional saccadic model.
\newblock {\em Neurocomputing}, 404:154 -- 164, 2020.

\bibitem{blondel2020differentiable}
Mathieu Blondel, Arthur Mensch, and Jean-Philippe Vert.
\newblock Differentiable divergences between time series.
\newblock {\em arXiv preprint arXiv:2010.08354}, 2020.

\bibitem{borji2012cvpr}
A. {Borji}.
\newblock Boosting bottom-up and top-down visual features for saliency
  estimation.
\newblock In {\em 2012 IEEE Conference on Computer Vision and Pattern
  Recognition}, 2012.

\bibitem{mit-saliency-benchmark}
Zoya Bylinskii, Tilke Judd, Ali Borji, Laurent Itti, Fr{\'e}do Durand, Aude
  Oliva, and Antonio Torralba.
\newblock Mit saliency benchmark.
\newblock http://saliency.mit.edu/, 2019.

\bibitem{cao2014look}
Ying Cao, Rynson~WH Lau, and Antoni~B Chan.
\newblock Look over here: Attention-directing composition of manga elements.
\newblock {\em ACM Trans. Graph.}, 33(4):1--11, 2014.

\bibitem{Chang_2019_CVPR}
Chien-Yi Chang, De-An Huang, Yanan Sui, Li Fei-Fei, and Juan~Carlos Niebles.
\newblock D3tw: Discriminative differentiable dynamic time warping for weakly
  supervised action alignment and segmentation.
\newblock In {\em Proceedings of the IEEE/CVF Conference on Computer Vision and
  Pattern Recognition (CVPR)}, June 2019.

\bibitem{chao2018salgan360}
Fang-Yi Chao, Lu Zhang, Wassim Hamidouche, and Olivier Deforges.
\newblock Salgan360: Visual saliency prediction on 360 degree images with
  generative adversarial networks.
\newblock In {\em 2018 IEEE Int. Conf. on Multim. \& Expo Workshops (ICMEW)},
  pages 01--04. IEEE, 2018.

\bibitem{colburn2000role}
Alex Colburn, Michael~F Cohen, and Steven Drucker.
\newblock The role of eye gaze in avatar mediated conversational interfaces.
\newblock Technical report, Citeseer, 2000.

\bibitem{coors2018spherenet}
Benjamin Coors, Alexandru Paul~Condurache, and Andreas Geiger.
\newblock Spherenet: Learning spherical representations for detection and
  classification in omnidirectional images.
\newblock In {\em Proc. of the European Conference on Computer Vision (ECCV)},
  pages 518--533, 2018.

\bibitem{cornia2018predicting}
Marcella Cornia, Lorenzo Baraldi, Giuseppe Serra, and Rita Cucchiara.
\newblock Predicting human eye fixations via an lstm-based saliency attentive
  model.
\newblock {\em IEEE Transactions on Image Processing}, 27(10):5142--5154, 2018.

\bibitem{cuturi2017soft}
Marco Cuturi and Mathieu Blondel.
\newblock Soft-dtw: a differentiable loss function for time-series.
\newblock {\em arXiv preprint arXiv:1703.01541}, 2017.

\bibitem{ellis1985patterns}
Stephen~R Ellis and James~Darrell Smith.
\newblock Patterns of statistical dependency in visual scanning.
\newblock {\em Eye movements and human information processing}, pages 221--238,
  1985.

\bibitem{fahimi2020metrics}
Ramin Fahimi and Neil~DB Bruce.
\newblock On metrics for measuring scanpath similarity.
\newblock {\em Behavior Research Methods}, pages 1--20, 2020.

\bibitem{horley2004face}
Kaye Horley, Leanne~M Williams, Craig Gonsalvez, and Evian Gordon.
\newblock Face to face: visual scanpath evidence for abnormal processing of
  facial expressions in social phobia.
\newblock {\em Psychiatry research}, 127(1-2):43--53, 2004.

\bibitem{itti1998model}
Laurent Itti, Christof Koch, and Ernst Niebur.
\newblock A model of saliency-based visual attention for rapid scene analysis.
\newblock {\em IEEE Transactions on pattern analysis and machine intelligence},
  20(11):1254--1259, 1998.

\bibitem{judd2009learning}
Tilke Judd, Krista Ehinger, Fr{\'e}do Durand, and Antonio Torralba.
\newblock Learning to predict where humans look.
\newblock In {\em IEEE ICCV}, pages 2106--2113. IEEE, 2009.

\bibitem{Adam}
Diederik~P. Kingma and Jimmy Ba.
\newblock Adam: A method for stochastic optimization.
\newblock In {\em ICLR}, 2014.
\newblock Last updated in arXiv in 2017.

\bibitem{kummerer2016deepgaze}
Matthias Kümmerer, Thomas S.~A. Wallis, and Matthias Bethge.
\newblock Deepgaze ii: Reading fixations from deep features trained on object
  recognition.
\newblock {\em arXiv preprint arXiv:1610.01563}, 2016.

\bibitem{lemeur2013}
O. Le~Meur and T. Baccino.
\newblock Methods for comparing scanpaths and saliency maps: strengths and
  weaknesses.
\newblock {\em Behavior Research Methods}, pages 251--266, 2013.

\bibitem{lemeur2015saccadic}
Olivier {Le Meur} and Zhi Liu.
\newblock Saccadic model of eye movements for free-viewing condition.
\newblock {\em Vision Research}, 116:152 -- 164, 2015.

\bibitem{li2019very}
Chenge Li, Weixi Zhang, Yong Liu, and Yao Wang.
\newblock Very long term field of view prediction for 360-degree video
  streaming.
\newblock In {\em 2019 IEEE Conference on Multimedia Information Processing and
  Retrieval (MIPR)}, pages 297--302. IEEE, 2019.

\bibitem{ling2019prediction}
Suiyi Ling, Jes{\'u}s Guti{\'e}rrez, Ke Gu, and Patrick Le~Callet.
\newblock Prediction of the influence of navigation scan-path on perceived
  quality of free-viewpoint videos.
\newblock {\em IEEE Journal on Emerging and Sel. Topics in Circ. and Sys.},
  9(1):204--216, 2019.

\bibitem{liu2013semantically}
Huiying Liu, Dong Xu, Qingming Huang, Wen Li, Min Xu, and Stephen Lin.
\newblock Semantically-based human scanpath estimation with hmms.
\newblock In {\em Proceedings of the IEEE International Conference on Computer
  Vision}, pages 3232--3239, 2013.

\bibitem{liu2018intriguing}
Rosanne Liu, Joel Lehman, Piero Molino, Felipe~Petroski Such, Eric Frank, Alex
  Sergeev, and Jason Yosinski.
\newblock An intriguing failing of convolutional neural networks and the
  coordconv solution.
\newblock In {\em Neural information processing systems}, pages 9605--9616,
  2018.

\bibitem{lu2012cvpr}
Y. {Lu}, W. {Zhang}, C. {Jin}, and X. {Xue}.
\newblock Learning attention map from images.
\newblock In {\em 2012 IEEE Conference on Computer Vision and Pattern
  Recognition}, 2012.

\bibitem{martin2021multimodality}
Daniel Martin, Sandra Malpica, Diego Gutierrez, Belen Masia, and Ana Serrano.
\newblock Multimodality in {VR}: A survey.
\newblock {\em arXiv preprint arXiv:2101.07906}, 2021.

\bibitem{martin20saliency}
Daniel Martin, Ana Serrano, and Belen Masia.
\newblock Panoramic convolutions for $360^{\circ}$ single-image saliency
  prediction.
\newblock In {\em CVPR Workshop on CV for AR/VR}, 2020.

\bibitem{mirza2014conditional}
Mehdi Mirza and Simon Osindero.
\newblock Conditional generative adversarial nets.
\newblock {\em arXiv preprint arXiv:1411.1784}, 2014.

\bibitem{monroy2018salnet}
Rafael Monroy, Sebastian Lutz, Tejo Chalasani, and Aljosa Smolic.
\newblock Salnet360: Saliency maps for omni-directional images with cnn.
\newblock {\em Signal Processing: Image Communication}, 69:26 -- 34, 2018.

\bibitem{muller2007dynamic}
Meinard M{\"u}ller.
\newblock Dynamic time warping.
\newblock {\em Information retrieval for music and motion}, pages 69--84, 2007.

\bibitem{nguyen2018your}
Anh Nguyen, Zhisheng Yan, and Klara Nahrstedt.
\newblock Your attention is unique: Detecting 360-degree video saliency in
  head-mounted display for head movement prediction.
\newblock In {\em Proc. ACM Intern. Conf. on Multimedia}, pages 1190--1198,
  2018.

\bibitem{Pan_2017_SalGAN}
Junting Pan, Cristian Canton, Kevin McGuinness, Noel~E. O'Connor, Jordi Torres,
  Elisa Sayrol, and Xavier~and Giro-i Nieto.
\newblock Salgan: Visual saliency prediction with generative adversarial
  networks.
\newblock 2018.

\bibitem{Pan_2016_CVPR}
Junting Pan, Elisa Sayrol, Xavier Giro-i Nieto, Kevin McGuinness, and Noel~E.
  O'Connor.
\newblock Shallow and deep convolutional networks for saliency prediction.
\newblock In {\em The IEEE Conference on Computer Vision and Pattern
  Recognition (CVPR)}, June 2016.

\bibitem{pang2016directing}
Xufang Pang, Ying Cao, Rynson~WH Lau, and Antoni~B Chan.
\newblock Directing user attention via visual flow on web designs.
\newblock {\em ACM Trans. on Graph.}, 35(6):1--11, 2016.

\bibitem{rai2017dataset}
Yashas Rai, Jes{\'u}s Guti{\'e}rrez, and Patrick Le~Callet.
\newblock A dataset of head and eye movements for 360 degree images.
\newblock In {\em Proceedings of the 8th ACM on Multimedia Systems Conference},
  pages 205--210, 2017.

\bibitem{ruhland2015review}
Kerstin Ruhland, Christopher~E Peters, Sean Andrist, Jeremy~B Badler, Norman~I
  Badler, Michael Gleicher, Bilge Mutlu, and Rachel McDonnell.
\newblock A review of eye gaze in virtual agents, social robotics and hci:
  Behaviour generation, user interaction and perception.
\newblock In {\em Computer graphics forum}, volume~34, pages 299--326. Wiley
  Online Library, 2015.

\bibitem{sela2017gazegan}
Matan Sela, Pingmei Xu, Junfeng He, Vidhya Navalpakkam, and Dmitry Lagun.
\newblock Gazegan-unpaired adversarial image generation for gaze estimation.
\newblock {\em arXiv preprint arXiv:1711.09767}, 2017.

\bibitem{Serrano_VR-cine_SIGGRAPH2017}
Ana Serrano, Vincent Sitzmann, Jaime Ruiz-Borau, Gordon Wetzstein, Diego
  Gutierrez, and Belen Masia.
\newblock Movie editing and cognitive event segmentation in virtual reality
  video.
\newblock {\em ACM Trans. Graph. (SIGGRAPH)}, 36(4), 2017.

\bibitem{sitzmann2018saliency}
Vincent Sitzmann, Ana Serrano, Amy Pavel, Maneesh Agrawala, Diego Gutierrez,
  Belen Masia, and Gordon Wetzstein.
\newblock Saliency in {VR}: How do people explore virtual environments?
\newblock {\em IEEE Trans. on Vis. and Comp. Graph.}, 24(4):1633--1642, 2018.

\bibitem{startsev2018360}
Mikhail Startsev and Michael Dorr.
\newblock 360-aware saliency estimation with conventional image saliency
  predictors.
\newblock {\em Signal Proces.: Image Comm.}, 69:43--52, 2018.

\bibitem{su2017making}
Yu-Chuan Su and Kristen Grauman.
\newblock Making 360 video watchable in 2d: Learning videography for click free
  viewing.
\newblock In {\em 2017 IEEE Conference on Computer Vision and Pattern
  Recognition (CVPR)}, pages 1368--1376. IEEE, 2017.

\bibitem{su2016pano2vid}
Yu-Chuan Su, Dinesh Jayaraman, and Kristen Grauman.
\newblock Pano2vid: Automatic cinematography for watching 360$^{\circ}$ videos.
\newblock In {\em Asian Conf. on CV}, pages 154--171. Springer, 2016.

\bibitem{tatler2009prominence}
Benjamin~W Tatler and Benjamin~T Vincent.
\newblock The prominence of behavioural biases in eye guidance.
\newblock {\em Visual Cognition}, 17(6-7):1029--1054, 2009.

\bibitem{tavakoli2013stochastic}
Hamed~Rezazadegan Tavakoli, Esa Rahtu, and Janne Heikkil{\"a}.
\newblock Stochastic bottom--up fixation prediction and saccade generation.
\newblock {\em Image and Vision Computing}, 31(9):686--693, 2013.

\bibitem{torralba2006contextual}
Antonio Torralba, Aude Oliva, Monica~S Castelhano, and John~M Henderson.
\newblock Contextual guidance of eye movements and attention in real-world
  scenes: the role of global features in object search.
\newblock {\em Psychological review}, 113(4):766, 2006.

\bibitem{Vig_2014_CVPR}
Eleonora Vig, Michael Dorr, and David Cox.
\newblock Large-scale optimization of hierarchical features for saliency
  prediction in natural images.
\newblock In {\em Proceedings of the IEEE Conference on Computer Vision and
  Pattern Recognition (CVPR)}, June 2014.

\bibitem{vincent2019shape}
LE Vincent and Nicolas Thome.
\newblock Shape and time distortion loss for training deep time series
  forecasting models.
\newblock In {\em Advances in neural information processing systems}, pages
  4189--4201, 2019.

\bibitem{saliencytoolbox}
Dirk Walther and Christof Koch.
\newblock Modeling attention to salient proto-objects.
\newblock {\em Neural Networks}, 19:1395--1407, 2006.

\bibitem{wang2017deep}
Wenguan Wang and Jianbing Shen.
\newblock Deep visual attention prediction.
\newblock {\em IEEE Transactions on Image Processing}, 27(5):2368--2378, 2017.

\bibitem{wang2018attention}
W. {Wang} and J. {Shen}.
\newblock Deep visual attention prediction.
\newblock {\em IEEE Transactions on Image Processing}, 27(5):2368--2378, 2018.

\bibitem{Wang_2018_CVPR}
Wenguan Wang, Jianbing Shen, Xingping Dong, and Ali Borji.
\newblock Salient object detection driven by fixation prediction.
\newblock In {\em Proceedings of the IEEE Conference on Computer Vision and
  Pattern Recognition (CVPR)}, June 2018.

\bibitem{wu2020spherical}
Chenglei Wu, Ruixiao Zhang, Zhi Wang, and Lifeng Sun.
\newblock A spherical convolution approach for learning long term viewport
  prediction in 360 immersive video.
\newblock In {\em Proceedings of the AAAI Conference on Artificial
  Intelligence}, volume~34, pages 14003--14040, 2020.

\bibitem{xia2019predicting}
Chen Xia, Junwei Han, Fei Qi, and Guangming Shi.
\newblock Predicting human saccadic scanpaths based on iterative representation
  learning.
\newblock {\em IEEE Transactions on Image Processing}, 28(7):3502--3515, 2019.

\bibitem{xu2019pami}
M. {Xu}, Y. {Song}, J. {Wang}, M. {Qiao}, L. {Huo}, and Z. {Wang}.
\newblock Predicting head movement in panoramic video: A deep reinforcement
  learning approach.
\newblock {\em IEEE Transactions on Pattern Analysis and Machine Intelligence},
  41(11):2693--2708, 2019.

\bibitem{yang2013saliency}
Chuan Yang, Lihe Zhang, Ruan Lu, Huchuan, Xiang, and Ming-Hsuan Yang.
\newblock Saliency detection via graph-based manifold ranking.
\newblock In {\em Computer Vision and Pattern Recognition (CVPR), 2013 IEEE
  Conference on}, pages 3166--3173. IEEE, 2013.

\bibitem{yun2013exploring}
Kiwon Yun, Yifan Peng, Dimitris Samaras, Gregory~J Zelinsky, and Tamara~L Berg.
\newblock Exploring the role of gaze behavior and object detection in scene
  understanding.
\newblock {\em Frontiers in psychology}, 4:917, 2013.

\bibitem{zhao2011saliency}
Qi Zhao and Christof Koch.
\newblock Learning a saliency map using fixated locations in natural scenes.
\newblock {\em Journal of Vision}, 11:9, 2011.

\bibitem{zhu2018prediction}
Yucheng Zhu, Guangtao Zhai, and Xiongkuo Min.
\newblock The prediction of head and eye movement for 360 degree images.
\newblock {\em Signal Processing: Image Communication}, 69:15--25, 2018.

\end{thebibliography}
}

\end{document}